\documentclass[11pt]{article}

\usepackage[final]{acl}
\usepackage{times}
\usepackage{latexsym}
\usepackage[T1]{fontenc}
\usepackage[utf8]{inputenc}
\usepackage{microtype}
\usepackage{courier}
\usepackage{graphicx}
\usepackage{amsmath}
\usepackage{amssymb}
\usepackage{booktabs}
\usepackage{array}
\newcolumntype{P}[1]{>{\raggedright\arraybackslash}p{#1}}
\usepackage{multirow}
\usepackage{stfloats}
\usepackage{xcolor}
\usepackage{colortbl}
\definecolor{palB}{HTML}{3B75AF}\definecolor{palQ}{HTML}{A0647D}\definecolor{palQP}{HTML}{B08A3E}
\definecolor{palRescue}{HTML}{3B75AF}\definecolor{palDamage}{HTML}{C4783B}
\usepackage{xurl}
\usepackage{enumitem}
\usepackage{tikz}
\usetikzlibrary{positioning,arrows.meta,fit,shapes.geometric}

\newcommand{\tautwo}{$\tau^2$-bench}
\newcommand{\pp}{\,pp}
\newcommand{\YY}[1]{Y_{\mathrm{#1}}}
\newcommand{\Gall}{G_{\mathrm{all}}}
\newcommand{\DQ}{\Delta_{\mathrm{Q}}}
\newcommand{\DB}{\Delta_{\mathrm{B}}}

\title{Measuring the Task Value of Precision Upgrades in Closed-Loop Agents}

\author{
  \textbf{Sicheng Lyu}$^{1,*}$ \quad \textbf{Xinyu Wang}$^{1,*}$ \quad \textbf{Yu Gu}$^{1}$ \\
  \textbf{Lei Ding}$^{3}$ \quad \textbf{Ziyang Song}$^{2}$ \quad \textbf{Xiao-Wen Chang}$^{1,\dagger}$ \\
  {$^{1}$McGill University \quad $^{2}$Ohio University \quad $^{3}$University of Manitoba} \\
  {\texttt{\{sicheng.lyu,xinyu.wang5\}@mail.mcgill.ca}} \\
  {\texttt{chang@cs.mcgill.ca}}
}

\begin{document}
\raggedbottom
\maketitle
\begingroup
\renewcommand{\thefootnote}{\fnsymbol{footnote}}
\footnotetext[1]{Equal contribution.}
\footnotetext[2]{Corresponding author.}
\endgroup

\begin{abstract}
A low-precision agent can call its full-precision (BF16) model at selected decisions as a fallback. A learned precision router chooses where to call it from the disagreement between the two models and is judged by end-to-end success. An inserted BF16 action, however, can rescue a failing task or damage a succeeding one. We argue that a fallback's value is the number of tasks it saves minus the number it spoils, measured for the intervention that will actually run. From restored states, we run paired branches that cross which model supplies the current action with which model continues, separating replacement of one action, takeover of later execution, and both. On Qwen3 agents with simulated 4-bit weights in \tautwo{}, replacing the first action where the two models differed in tool use changed many task outcomes, but rescues and damages nearly cancelled, and on held-out telecom tasks the action-only estimate was negative. Replacing both the action and the continuation had a positive net gain in two of five exploratory lower-bit stress settings. Two references changed outcomes at fixed bit width: replacing an action with that of a second 4-bit checkpoint changed about as many outcomes as replacing it with the BF16 action, and reordering tool-argument keys changed outcomes while the executed tool call stayed identical. At common states, the additional value of the BF16 action over that checkpoint remained unresolved. A frozen uncertainty trigger found more outcome changes than random selection, while its net benefit stayed within the range of random selection. We recommend evaluating each fallback by its signed task benefit under the intended continuation, against same-precision references, before attributing gains to more bits.
Our code is available at \url{https://github.com/W4ng77/cloq}.
\end{abstract}

\section{Introduction}
\label{sec:intro}

Weight quantization \citep{frantar2023gptq,lin2024awq} and key-value (KV) cache quantization \citep{liu2024kivi} make large language model (LLM) agents cheaper to serve, and a quantized agent can keep its full-precision model in reserve. Precision fallback calls the BF16 model at selected decisions of an interaction \citep{li2026dynamicmp}. The BF16 model may replace the current action, take over from the next step, or do both, and each choice changes a different part of the running trajectory.

The precision router of \citet{li2026dynamicmp} learns where to fall back from the Kullback--Leibler (KL) divergence between the two models' action distributions and is evaluated by end-to-end task success. Both quantities measure something other than the value of one fallback. An end-to-end score describes a model running from the start, while a fallback inserts that model into a history built by another. Disagreement describes how two models differ. Compressed models already change answers in both directions on static tasks \citep{dutta2024accuracy} and on agent tasks \citep{jang2026flat}, and a changed agent action can rescue a failure, damage a success, or leave the outcome as it was. The value of a fallback is the number of rescues minus the number of damages.

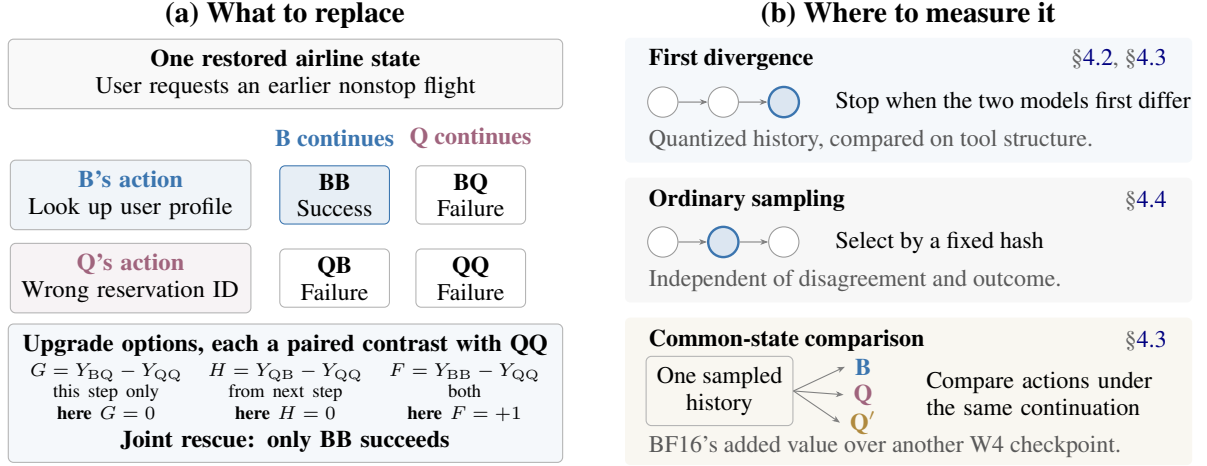
\begin{figure*}[t]
\centering
\begin{minipage}[t]{.49\textwidth}
\centering
\textbf{(a) What to replace}\par\smallskip
\begin{tikzpicture}[ %
  font=\footnotesize, %
  box/.style={draw=black!30,rounded corners=2pt,align=center,inner sep=4pt}, %
  arm/.style={box,inner sep=2pt,minimum width=1.45cm,minimum height=.77cm}, %
  arr/.style={-{Stealth[length=4pt]},draw=black!45}] %
\node[box,text width=7.05cm,fill=gray!5] (s) at (0,0) %
  {\textbf{One restored airline state}\\User requests an earlier nonstop flight};
\node[font=\footnotesize\bfseries,text=palB] at (.65,-.85) {B continues}; %
\node[font=\footnotesize\bfseries,text=palQ] at (2.45,-.85) {Q continues}; %
\node[box,text width=2.9cm,fill=palB!7,anchor=east] (ab) at (-.45,-1.6) %
  {\textcolor{palB}{\textbf{B's action}}\\Look up user profile};
\node[box,text width=2.9cm,fill=palQ!8,anchor=east] (aq) at (-.45,-2.7) %
  {\textcolor{palQ}{\textbf{Q's action}}\\Wrong reservation ID};
\node[arm,fill=palRescue!12,draw=palRescue] (bb) at (.65,-1.6) {\textbf{BB}\\Success}; %
\node[arm] (bq) at (2.45,-1.6) {\textbf{BQ}\\Failure}; %
\node[arm] (qb) at (.65,-2.7) {\textbf{QB}\\Failure}; %
\node[arm] (qq) at (2.45,-2.7) {\textbf{QQ}\\Failure}; %
\node[box,text width=7.05cm,fill=palRescue!5,anchor=north] at (0,-3.3) %
  {\textbf{Upgrade options, each a paired contrast with QQ}\\[1pt]{\scriptsize\setlength{\tabcolsep}{1pt}\begin{tabular}{@{}*{3}{>{\centering\arraybackslash}p{2.3cm}}@{}}$G=\YY{BQ}-\YY{QQ}$ & $H=\YY{QB}-\YY{QQ}$ & $F=\YY{BB}-\YY{QQ}$\\ this step only & from next step & both\\ \textbf{here} $G=0$ & \textbf{here} $H=0$ & \textbf{here} $F=+1$\end{tabular}}\\[1pt]\textbf{Joint rescue: only BB succeeds}}; %
\end{tikzpicture}
\end{minipage}\hfill
\begin{minipage}[t]{.48\textwidth}
\centering
\textbf{(b) Where to measure it}\par\smallskip
\begin{tikzpicture}[x=1cm,y=1cm,font=\footnotesize, %
 state/.style={circle,draw=black!40,fill=white,minimum size=4mm,inner sep=0pt}, %
 selected/.style={state,draw=palB,fill=palB!18,line width=.9pt}, %
 flow/.style={-{Stealth[length=3pt]},draw=black!50}, %
 head/.style={anchor=west,font=\footnotesize\bfseries},
 note/.style={anchor=west,align=left,font=\footnotesize}]
\useasboundingbox (0,.3) rectangle (7.5,-5.35); %
\fill[palB!5,rounded corners=2pt] (0,.3) rectangle (7.5,-1.35); %
\node[head] at (.18,.02) {First divergence}; %
\node[anchor=east,text=black!60] at (7.3,.02) {\S\ref{sec:value}, \S\ref{sec:sameprec}}; %
\node[state] (a1) at (.5,-.55) {}; %
\node[state] (a2) at (1.3,-.55) {}; %
\node[selected] (a3) at (2.1,-.55) {}; %
\draw[flow] (a1)--(a2); \draw[flow] (a2)--(a3); %
\node[note] at (2.65,-.55) {Stop when the two models first differ}; %
\node[note,text=black!65] at (.18,-1.06) {Quantized history, compared on tool structure.}; %
\fill[black!3,rounded corners=2pt] (0,-1.55) rectangle (7.5,-3.2); %
\node[head] at (.18,-1.85) {Ordinary sampling}; %
\node[anchor=east,text=black!60] at (7.3,-1.85) {\S\ref{sec:trigger}}; %
\node[state] (b1) at (.5,-2.4) {}; %
\node[selected] (b2) at (1.3,-2.4) {}; %
\node[state] (b3) at (2.1,-2.4) {}; %
\draw[flow] (b1)--(b2); \draw[flow] (b2)--(b3); %
\node[note] at (2.65,-2.4) {Select by a fixed hash}; %
\node[note,text=black!65] at (.18,-2.91) {Independent of disagreement and outcome.}; %
\fill[palQP!7,rounded corners=2pt] (0,-3.4) rectangle (7.5,-5.35); %
\node[head] at (.18,-3.7) {Common-state comparison}; %
\node[anchor=east,text=black!60] at (7.3,-3.7) {\S\ref{sec:sameprec}}; %
\node[draw=black!30,rounded corners=2pt,align=center,inner sep=4pt] (h) at (1.25,-4.37) {One sampled\\history}; %
\node[text=palB,font=\footnotesize\bfseries] (c1) at (3.15,-4.05) {B}; %
\node[text=palQ,font=\footnotesize\bfseries] (c2) at (3.15,-4.42) {Q}; %
\node[text=palQP,font=\footnotesize\bfseries] (c3) at (3.15,-4.79) {Q$'$}; %
\draw[flow] (h.east)--(c1.west);\draw[flow] (h.east)--(c2.west);\draw[flow] (h.east)--(c3.west); %
\node[note,text width=3.2cm] at (3.85,-4.4) {Compare actions under\\the same continuation}; %
\node[note,text=black!65] at (.18,-5.1) {BF16's added value over another W4 checkpoint.}; %
\end{tikzpicture}

\end{minipage}
\caption{Fallback interventions and state sources. Q is the quantized model and B its BF16 version. Branch XY lets X supply the current action and Y continue, and $\YY{XY}$ is its task success. (a) Four branches from one restored airline state (Qwen3-4B with 4-bit weights quantized by AWQ-style scaling, task 31, decision 1 of 12), crossing the action source (rows) with the continuing model (columns). The box below defines the upgrades $G$, $H$ and $F$ against QQ and gives their values here. (b) Schematic of the three state sources defined in Section~\ref{sec:states}.} %
\label{fig:case}
\label{fig:fork}
\label{fig:method}
\end{figure*}

We measure this value for each fallback mode at a single decision. From a restored state, we execute the actions of the quantized model and its BF16 version and continue each with either model, which gives four complete branches (Figure~\ref{fig:case}a). Each branch is named by its action source and continuing model, with B for the BF16 model and Q for the quantized model, so BB uses the BF16 model throughout and QQ is the quantized baseline. Their contrasts with QQ give the value of replacing the action, taking over later execution, and both. These contrasts extend step values under a continuing policy \citep{setlur2024rewarding} and per-step model swaps \citep{gonuguntla2026replay}. In the airline example of Figure~\ref{fig:case}a, the quantized model looks up a reservation id that does not exist and the BF16 model looks up the user's profile. Only BB succeeds, so the BF16 action helped only when the BF16 model also continued, and an action-only evaluation would have missed this rescue. Across Qwen3 agents with 4-bit weights (W4), replacing one action changed many outcomes in both directions and left average success nearly unchanged. %

A changed outcome also leaves open whether the change required more bits. We add two references at fixed bit width: a second 4-bit checkpoint built from resampled calibration data, and a reordering of the keys in the quantized model's tool arguments, which changes the token sequence while the executed operation stays identical. We study Qwen3-4B and Qwen3-8B \citep{yang2025qwen3} under simulated weight and KV-cache quantization in \tautwo{} \citep{yao2024tau,barres2025tau2}, with a pre-registered holdout of new telecom tasks. Scope checks cover BFCL multi-turn \citep{patil2025bfcl}, ALFWorld \citep{shridhar2021alfworld} and Mistral-Small-3.2-24B \citep{mistral2025small32} (Appendix~\ref{app:scope}). %

Three findings build the argument:
\begin{enumerate}[leftmargin=*,itemsep=2pt,topsep=3pt]
\item \textbf{A fallback saves some tasks and spoils others.} Action replacement changed outcomes in both directions with little pooled net gain, and on the held-out telecom tasks its estimate was negative. Replacing both the action and the continuation had a positive net gain in two exploratory lower-bit stress settings (\S\ref{sec:value}).
\item \textbf{Outcome changes also occur at fixed bit width.} The second 4-bit checkpoint and the key reordering both changed outcomes, and the additional value of the BF16 action at common states remained unresolved (\S\ref{sec:sameprec}). %
\item \textbf{Finding sensitive decisions and selecting beneficial ones are separate tests.} A frozen uncertainty trigger found more outcome changes than random selection, while its net benefit stayed within the range of random selection (\S\ref{sec:r4}).
\end{enumerate}

\section{Related Work}
\label{sec:related}

\paragraph{Quantized agents compared episode by episode.} \emph{Flat Score, Amplified Failures} \citep{jang2026flat} is closest in setting: on \tautwo{} retail and telecom, 4-bit agents keep a flat task score in all ten model-domain cells while tool-call errors rise significantly in two, and in its highest-damage cell paired tasks flip in both directions. %
Because its episodes are sampled independently, the failing action and the recovering controller change precision together, whereas we vary them separately at a fixed state. Two-sided flips under compression were shown on static tasks \citep{dutta2024accuracy}, and compression studies compare formats, tasks and models by aggregate scores \citep{kurtic2025bf16,dong2025acbench}.

\paragraph{Forks and counterfactual replay.} \emph{The Replay Gap} \citep{gonuguntla2026replay} forks live SWE-bench Verified trajectories for per-step model switching. Its swap replaces the next action and continuation together, corresponding to our joint intervention. Its control holds the model fixed, and ours holds the bit width fixed. %
Decisive errors are defined as steps whose correction would turn a failure into a success \citep{zhang2025whichagent}. Replacing one action at a restored state and letting the original policy continue (our ``this step only'' contrast) locates them \citep{zhang2025agentracer}, attributes failures \citep{shah2026car} and yields training signal \citep{li2026cso,li2026crr}. %

\paragraph{Precision routing, escalation and their signals.} \citet{li2026dynamicmp} learn a per-step router between BF16 and 4-bit weights from thresholded KL divergence labels, then refine it with a reward that combines task success and latency. Sensitivity has been localised to phases \citep{lu2026mixquant}, task types \citep{zhang2026quantclaw} and high-entropy, high-KL tokens \citep{lotfi2026quantized}. Agents also defer single decisions when uncertain \citep{piatrashyn2026redact}, learn cost-aware escalation \citep{lei2026tacit}, or are routed per step against execution-verified labels \citep{yang2026twinrouterbench}. Continuing another model's trajectory has its own cost \citep{ganz2026handoff}. Under executed replay, confidence located consequential steps poorly \citep{zhang2026credit}, and KL measures the size of a disagreement while leaving its direction open \citep{nikolic2026displacement}. These works choose where to escalate. We measure the signed task value of an escalation once it is chosen. %

\paragraph{Evaluation noise and user simulators.} Prefix caching changed multi-turn tool-use trajectories in 36.2\% of episodes at 16-bit and 75.0\% at 4-bit \citep{patodiya2026same}, and agentic evaluations vary from run to run \citep{bjarnason2026randomness}, under semantics-preserving prompt changes \citep{chen2026noisefloor} and with the user model \citep{seshadri2026lost,jain2026userproxybench}. Our paired branches run on one fixed serving stack and user simulator. %

\section{Measuring Fallback Value}
\label{sec:method}

We evaluate a specified fallback at a recorded decision. We write Q for the quantized model and B for its BF16 version when naming branches and outcomes, and every compared branch starts from the same history and environment. Experiments repeat each task over prompt replicates called \emph{blocks}, and a \emph{unit} is one task in one block.

\subsection{Four branches on one state}
\label{sec:estimands}

\paragraph{Frozen state.} Before decision $t$ of the quantized model, we restore state $s_t$. It contains the full message history, the user simulator's visible conversation, and the used step and error budgets. Replaying the recorded tool calls rebuilds the environment and user-device databases. We save the state before running any branch, so all branches start from identical prompt tokens.

\paragraph{Four branches.} Let $\YY{XY}\in\{0,1\}$ denote task success when X supplies the current action and Y continues, with X, Y $\in\{\mathrm{B},\mathrm{Q}\}$. All branches start with the same remaining budget. Each upgrade is compared with QQ: replace this step ($G$), take over from the next step ($H$), or do both ($F$), as defined in Figure~\ref{fig:case}a. %
A \emph{rescue} occurs when the upgrade succeeds and QQ fails, and a \emph{damage} is the reverse. Either is an outcome \emph{flip}. Their sum is \emph{turnover} and their difference is \emph{net gain}. Additional notation appears in Appendix~\ref{app:names}.

\paragraph{The action being replaced.} An action is a complete assistant turn, including its language and structured tool calls. Its message and tool results enter the subsequent history, and the continuing model makes every later decision.  The four branches are complete task executions, and each task--block unit receives one intervention. A router that can trigger repeatedly is a different policy and needs an end-to-end evaluation.

\paragraph{Reference interventions.} Two references change the action while keeping the bit width fixed. The first repeats the design with a second checkpoint (Q$'$) in place of the BF16 model. It has the same bit width and is built by resampling states and activation rows from the same calibration trajectories as the quantized model. The second reverses the order of the keys in the quantized model's own tool arguments. The parsed call and its environment effects stay identical, and only the token sequence that the agent reads afterward changes. A third comparison substitutes the BF16 action and the second checkpoint's action at the same common states with the same continuing model, which measures the additional value of the BF16 action directly.

\subsection{State sources}
\label{sec:states}

A \emph{first divergent state} is the earliest decision in the quantized model's history where the two models differ in action kind, tool calls or normalised arguments. Differences in wording alone leave a state non-divergent. \emph{Ordinary states} are selected by a fixed hash before action generation, one per task and block, and a \emph{common state} is an ordinary state at which several action sources are compared (Figure~\ref{fig:method}b). Full-pool rates include every task--block unit and count units without a divergence as zero effect, while mode and pattern analyses condition on divergent states. We write $\Gall$ for the full-pool net gain of action replacement and $G$ for its average on divergent states. %

\section{Experiments} %
\label{sec:experiments}

\subsection{Experimental setup}
\label{sec:setting}

\paragraph{Tasks and agents.} We use \tautwo{} in text mode. The development panel has 58 retail and 27 airline tasks, and each W4 development configuration ran 20 prompt blocks. We also evaluate 80 telecom development tasks and a holdout of 34 telecom tasks whose protocol was fixed before any agent ran them. %
The agents are Qwen3-4B-Instruct and Qwen3-8B with thinking disabled. Mistral-Small-3.2-24B serves as a second family at ordinary states (Appendix~\ref{app:scope}). %
One greedy Qwen3-30B-A3B-Instruct user simulator serves all agents. %
Retail and airline success requires a final-database match and normal termination, and it excludes some official communication checks. Telecom uses the official task-basis success. BFCL multi-turn and ALFWorld have no LLM user, and each item is evaluated once (Appendix~\ref{app:setup}).

\paragraph{Quantization and serving.} The quantized model is a simulated (fake-quantized) checkpoint served with BF16 compute on the same path as the BF16 model. The main 4-bit weight (W4) settings use GPTQ \citep{frantar2023gptq} or AWQ-style scaling \citep{lin2024awq}, with group size 128 and a BF16 KV cache. Calibration uses BF16 agent trajectories from 57 retail and airline tasks, disjoint from the development panel. Separate KV8 configurations quantize the KV cache to 8-bit integers (INT8) while retaining BF16 weights. Lower-bit stress settings use 3-bit weights (W3), a rotated 4-bit KV cache (KV4-rot) or round-to-nearest quantization (RTN), and Appendix Table~\ref{tab:qconfig} lists every setting. %
Both models run on vLLM \citep{kwon2023vllm} with batch-invariant kernels \citep{he2025defeating} and deterministic compilation, in separate server processes without a shared prefix cache. A configuration is a model with a quantization setting. Each block runs the quantized model over the task pool with a fresh session-identifier line in the system prompt.

\paragraph{References.} The second-checkpoint reference covers Qwen3-4B and Qwen3-8B GPTQ-W4 on the development panel, with GPTQ-W3 as a stress check, and keeps the quantized episodes, tasks and prompt blocks fixed. Key reversal uses 300 eligible Qwen3-8B states from 41 tasks. The common-state comparison uses 510 ordinary states from Qwen3-8B BF16 histories. Each design was fixed before its first outcomes were read, and later amendments are listed in Appendix~\ref{app:timeline}. %

\paragraph{Replay audit.} Outcome-independent no-op audits checked that restoring a state preserves its original result. All 924 sampled development states and all 1202 holdout divergent states matched under the greedy, dense-BF16 serving stack. Appendix~\ref{app:setup} gives the acceptance rules and Appendix~\ref{app:prereg} the holdout protocol. %

\paragraph{Pooling and intervals.} Contrasts are paired within a configuration. The W4 pool averages four configuration-specific effects equally on the fixed development task set, and each configuration supplies its own first divergences, so the pool compares configurations at different decisions. Only complete blocks enter. %
Effects are in percentage points (pp). Main figures use new-task (crossed) intervals, which resample both tasks and prompt blocks, and the holdout uses its registered 90\% level. Fixed-pool intervals, which resample blocks over the fixed task set, appear in the appendix. An interval containing zero leaves an effect unresolved, and equivalence has its own test. Each branch measures one realised outcome under the greedy stack, and the stability of a rescue or damage across user or prompt conditions is a separate quantity. %

\subsection{Tasks saved and tasks spoiled}
\label{sec:value}
\label{sec:r1}
\label{sec:r2}

\begin{figure*}[t]
\centering
\includegraphics[width=\textwidth]{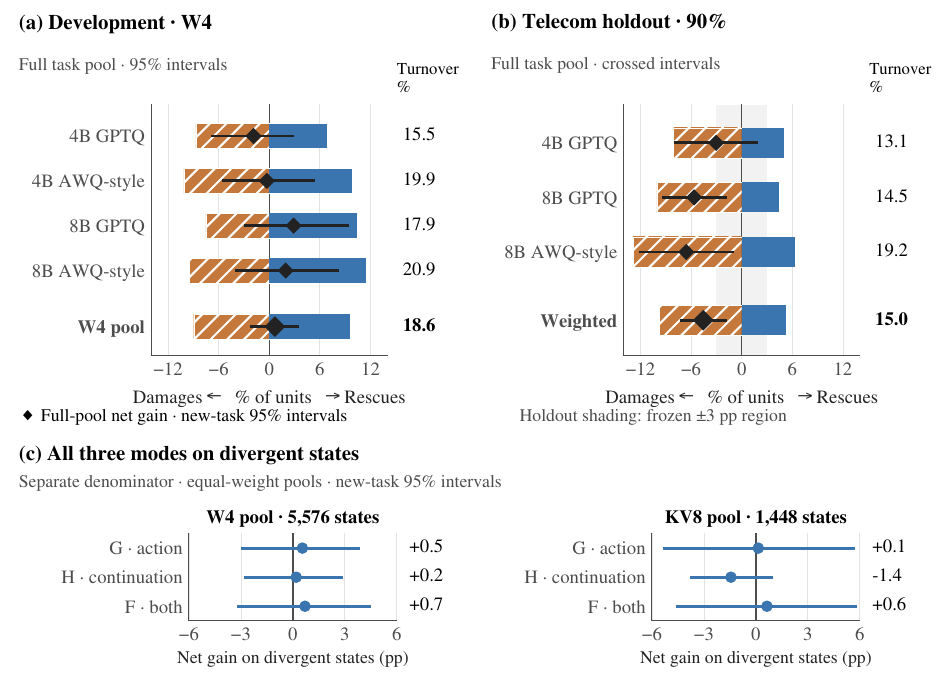}
\caption{Rescues, damages and net gain of action replacement. (a,b) Each row is one configuration or pool, in percentage of task--block units. Blue bars to the right show rescues, hatched orange bars to the left damages, and diamonds the full-pool net gain $\Gall$ with its interval. The right column gives turnover, rescues plus damages. (a) Development panel, new-task 95\% intervals. (b) Pre-registered telecom holdout, crossed 90\% intervals, with grey shading marking the registered $\pm$3\pp{} small-effect region. Bold rows are pools or the registered estimate. (c) Net gain on divergent states for action replacement ($G$), takeover ($H$) and joint replacement ($F$), with new-task 95\% intervals and the point estimate printed at the right. AWQ denotes AWQ-style scaling. Further intervals: Appendix~\ref{app:tables}.} %
\label{fig:rd}

\end{figure*}

\paragraph{Rescues and damages nearly balance.} At the first divergence on the development panel, the BF16 action replaced the quantized action and the quantized model continued. Outcomes changed in 15.5--20.9\% of task--block units across the four W4 configurations (Figure~\ref{fig:rd}a). Rescues and damages were of similar size, so the equally weighted full-pool net gain was close to zero with a new-task interval covering zero, and every configuration left its net gain on new tasks unresolved. %

\paragraph{The telecom holdout has a negative estimate.} The pre-registered holdout ran 34 unseen telecom tasks over 12 fresh blocks and three W4 configurations. Damages exceeded rescues, and the full-pool net gain had a crossed 90\% interval below zero (Figure~\ref{fig:rd}b). The registered readings had categories for positive and small effects and none for a negative effect, so the protocol classifies the result as \emph{inconclusive}. Both registered secondary checks passed: rescues and damages were both positive, and task fragility repeated across independent prompt blocks (Appendix~\ref{app:prereg}). %
Full baseline replay matched at every holdout divergent state, and the holdout changed domain as well as tasks (Appendix~\ref{app:holdoutdev}). %

In this hybrid trajectory the BF16 model supplies one action and the quantized model handles its consequences, and difficulty in continuing from the BF16 action is one candidate cause that these data leave untested.

\paragraph{The intervention defines which rescues count.} Figure~\ref{fig:rd}c reports all three modes on the W4 and KV8 divergent-state pools. For W4, action replacement, takeover and joint replacement had similar small averages, and all three new-task intervals cover zero.  On a few percent of W4 divergent states only changing both succeeded (\emph{joint rescue}), and on a similar share the BF16 action succeeded with the quantized model continuing but failed with the BF16 model continuing (\emph{rescue undone}). Against a label-permutation reference, joint rescue was above and rescue undone below the reference on the fixed task pool, and both new-task intervals cover zero (Appendix~\ref{app:offdiag}). An action-only evaluation misses joint rescues and a joint-replacement evaluation misses undone rescues, so each mode needs its own measurement.

\begin{table}[t]
\centering\small
\setlength{\tabcolsep}{3pt}\renewcommand{\arraystretch}{1.12} %
\begin{tabular*}{\columnwidth}{@{\extracolsep{\fill}}lrrr@{}}
\toprule
\textbf{Stress setting} & \textbf{$G$} & \textbf{$H$} & \textbf{$F$ [95\% CI]} \\
\midrule
4B GPTQ-W3 & $-1.3$ & $+4.3$ & \shortstack[r]{$+1.3$\\{}[$-6.8$, $+9.4$]} \\
4B KV4-rot & $+5.2$ & $+9.4$ & \shortstack[r]{\textbf{$+12.4$}\\{}[$+3.0$, $+21.6$]} \\
\addlinespace[2pt] %
8B GPTQ-W3 & $+2.1$ & $+0.4$ & \shortstack[r]{$+2.1$\\{}[$-6.6$, $+11.8$]} \\
8B KV4-rot & $+3.3$ & $-0.8$ & \shortstack[r]{$+4.6$\\{}[$-6.6$, $+15.9$]} \\
8B RTN-W3 & $-0.4$ & $+15.3$ & \shortstack[r]{\textbf{$+18.4$}\\{}[$+9.4$, $+28.2$]} \\
\bottomrule
\end{tabular*}
\caption{Net gain (pp) on divergent states in the five exploratory stress settings, Qwen3-4B (4B) and Qwen3-8B (8B) on retail and airline, three blocks each: action replacement $G$, takeover $H$ and joint replacement $F$. Only $F$ is shown with its unadjusted new-task 95\% confidence interval (CI), and bold marks an interval above zero. Intervals for $G$ and $H$: Table~\ref{tab:ghf}.}
\label{tab:scope}
\end{table}

The five lower-bit stress settings show larger effects (Table~\ref{tab:scope}). Joint replacement ($F$) had a positive new-task interval in Qwen3-4B KV4-rot and Qwen3-8B RTN-W3, and the other three settings stayed unresolved. In RTN-W3, borrowing one action had an estimate near zero while handing over later execution helped. 

\paragraph{Other environments, models and simulators.} On BFCL multi-turn and ALFWorld, action-only replacement also changed outcomes in both directions. Both W4 pools had net gains with intervals covering zero (Appendix Table~\ref{tab:settings}), one BFCL configuration had a negative net gain, and joint replacement had a positive net gain in two BFCL W4 configurations and in several lower-bit stress cells (Appendix~\ref{app:extr2}). For Mistral-Small-3.2-24B at its own ordinary states, the BF16 action rescued and damaged at similar rates and every mode's net value stayed unresolved (Appendix~\ref{app:mistral}). Individual labels were less stable than these averages. When a second user simulator replayed text-difference flips, most flips vanished or reversed, with the simulator model, sampling and relay interface changed together. A four-branch control on 80 states tracked the original per-state outcomes only moderately and had too few states to resolve within-configuration contrasts (Appendix~\ref{app:ordinary}). 

\subsection{Outcome changes at fixed bit width}
\label{sec:sameprec}
\label{sec:refs}

\begin{figure*}[t]
\centering
\includegraphics[width=\textwidth]{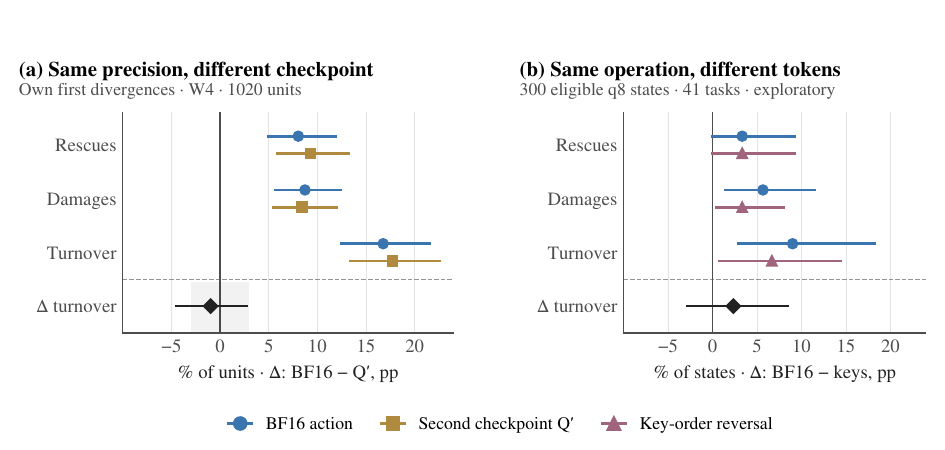}
\caption{Same-precision references. Rows give rescues, damages and turnover in percentage of units or states, with point estimates and 95\% intervals. Circles mark the BF16 action, squares the second W4 checkpoint (Q$'$) and triangles key-order reversal. Below the dashed line, $\Delta$ turnover is BF16 minus the reference in percentage points. (a) BF16 and the second checkpoint, each substituted at its own first divergence on 1020 task--block units, with new-task intervals. (b) BF16 action replacement and key-order reversal of the quantized tool call on the same 300 eligible states, with the wider of block and new-task intervals. The common-state comparison and all other intervals appear in Appendix~\ref{app:commonstate} and Table~\ref{tab:refsfull}.} %
\label{fig:refs}
\end{figure*}

A flip shows that an outcome is sensitive to the action at a decision. The two references test whether that sensitivity implies a need for more bits (Figure~\ref{fig:refs}). %

\paragraph{Resampling calibration changes outcomes at fixed bit width.} On the matched W4 task--block panel, substituting the second checkpoint at its own first divergence produced about as much turnover as substituting the BF16 model (Figure~\ref{fig:refs}a). Both substitutions rescued and damaged tasks. When the substitute also continued, the BF16 model changed more units than the second checkpoint (Appendix~\ref{app:qprime}). The turnover difference between them is unresolved, and its interval is too wide to establish equivalence within $\pm$3\pp{}. The W3 stress comparison also changed outcomes in both directions (Appendix~\ref{app:qprime}). %

\paragraph{Reordering argument keys keeps the tool call identical.} Key reversal alone altered outcomes at a lower point rate than BF16 action replacement on the same states, and their paired difference is unresolved (Figure~\ref{fig:refs}b). The experiment is exploratory and covers one representation change (Appendix~\ref{app:keyorder}). %

\begin{figure}[t]
\centering
\includegraphics[width=\columnwidth]{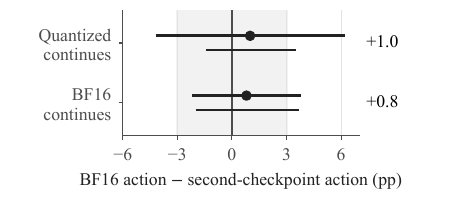}
\caption{Additional value of the BF16 action at 510 common Qwen3-8B states: success with the BF16 action minus success with the action of the second checkpoint (Q$'$), for each continuing model. Thick lines are 97\% $t$ intervals over block means and thin lines 97\% bootstrap intervals, both adjusted for the two looks described in Appendix~\ref{app:commonstate}, and the grey band is the $\pm$3\pp{} equivalence margin.} %
\label{fig:shared}
\end{figure}

\paragraph{The BF16 advantage needs common states.} The comparisons above substitute each model at its own first divergence, so the BF16 model and the second checkpoint act at different decisions. At the common states, the second checkpoint's action departed from the quantized action at least as often as the BF16 action did (Appendix~\ref{app:commonstate}). With the continuation held fixed, the additional value of the BF16 action was small and positive under either continuation (Figure~\ref{fig:shared}). Both adjusted intervals cover zero and extend past the $\pm$3\pp{} margin, so a useful BF16 advantage remains possible (Appendix~\ref{app:resolution}). Appendix~\ref{app:commonstate} gives the expansion and reading rules adopted after the first look. %

The references expose outcome changes that require neither more bits nor a different tool operation. We report them beside the BF16 turnover and leave them unsubtracted, since the two substitutes act at different states in Figure~\ref{fig:refs}a. 

\begin{figure*}[t]
\centering
\includegraphics[width=\textwidth]{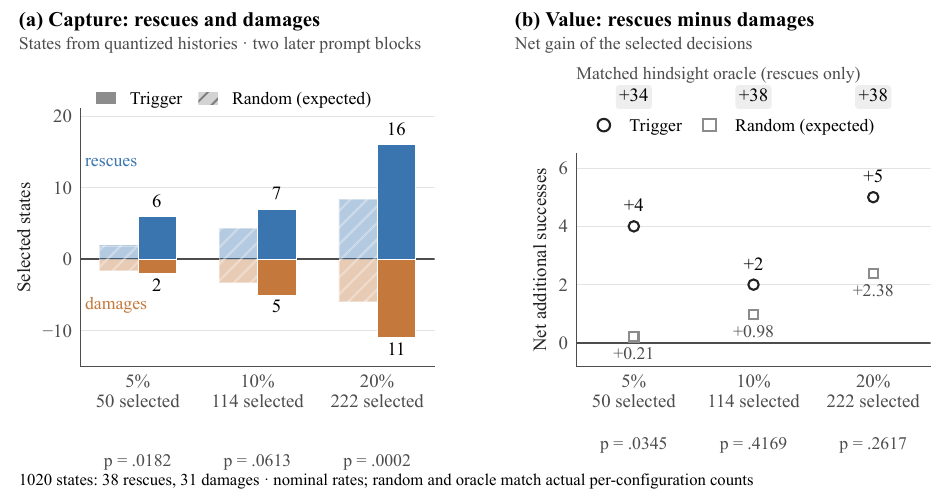}
\caption{Frozen margin trigger on 1020 validation states from quantized histories, by nominal rate with actual selected counts. (a) Rescues (up) and damages (down) for the trigger (solid) and expected under matched-count random selection (hatched). (b) Rescues minus damages for the trigger (circles) and random selection (squares). The grey strip gives the matched-count hindsight oracle, which selects rescues by construction. Unadjusted exact one-sided $p$-values compare each panel with random selection. Full values: Table~\ref{tab:oracle}.} %
\label{fig:trigger}
\label{fig:oracle}
\end{figure*}

\subsection{Selection for net benefit}
\label{sec:r4}
\label{sec:trigger}

A selector must find decisions where the fallback changes the outcome and, among those, decisions where it helps. We tested one frozen uncertainty rule on both counts (Figure~\ref{fig:trigger}).

\paragraph{A rule chosen before calling BF16.} The rule reads only the quantized model's output, so it can choose a decision before any BF16 computation. On one development block, we compared entropy, margin and mean log-probability by net gain over random selection and froze the winning margin rule before validation: when the minimum token margin, the smallest gap between the two most likely tokens over the quantized action, is low, the BF16 model supplies the action and the quantized model continues. This implements action replacement, and small top-two logit margins mark the tokens where greedy decoding diverges between BF16 and 16-bit floating-point (FP16) inference \citep{du2026greedy}. Per-configuration thresholds target 5, 10 and 20\% of development states. Validation can select a different fraction, so the random and oracle baselines match the actual counts. %

Validation used two later blocks of the same development tasks, separate from the telecom holdout, with states from the quantized model's own histories in four W4 and two KV8 configurations. The rule captured more outcome changes than matched random selection at every rate, with the clearest separation at the 20\% rate (Figure~\ref{fig:trigger}a). A secondary test on BF16 histories appears in Appendix~\ref{app:token}. %

\paragraph{Benefit against random selection.} The captured states included damages as well as rescues (Figure~\ref{fig:trigger}a). After Holm--Bonferroni correction across the three rates, which the protocol treated as co-equal, the net benefit over matched random selection stayed unresolved at every rate (Figure~\ref{fig:trigger}b). At the 5\% rate the unadjusted net test was nominally significant, and Appendix~\ref{app:oracle} gives all exact tests. The matched hindsight oracle selects rescues by construction and bounds what any selector could gain at these counts. %

This result concerns one rule on later blocks of the same tasks, and it motivates a signed-benefit test for learned routing and escalation \citep{li2026dynamicmp,piatrashyn2026redact,lei2026tacit}.

\section{Discussion and Conclusion}
\label{sec:discussion}
\label{sec:conclusion}
\label{sec:scope}

We have measured precision fallback by the tasks it saves and the tasks it spoils, separately for replacing one action, taking over later execution, and both. With 4-bit Qwen3 agents in \tautwo{}, action replacement has changed many outcomes in both directions with little net gain and has lowered success on new telecom tasks, while joint replacement has had positive net gains in two exploratory lower-bit stress settings. Outcome changes have also appeared at fixed bit width, and an uncertainty trigger that finds changed outcomes has left its net benefit unresolved. %

\paragraph{Measure the mode that will run.} A selector's training target should be the net gain of the mode it will execute, because a rescue under one mode can fail under another (Appendix~\ref{app:offdiag}). Measuring all three modes costs four complete task executions per state, against two for action replacement alone.

\paragraph{Keep the sign of the consequence.} Turnover counts a rescue and a damage alike, while task benefit gives them opposite signs. Disagreement or uncertainty is a cheap way to nominate decisions. The benefit of the nominated decisions should then be tested on held-out decisions against random selection with the same fallback budget. When modes are compared, that budget should count BF16 computation, since a takeover can use many more BF16 decisions than one borrowed action. We leave these costs unmeasured. 

\paragraph{Check precision with a reference.} Before attributing a flip to precision, compare the BF16 action with a same-bit-width reference at common states under the same continuation, at the cost of one extra checkpoint and two extra branches per state.

\paragraph{Relation to prior evidence.} The BF16 model is a reference policy whose advantage as an agent needs measuring, since Qwen3-4B GPTQ-W4 had a higher end-to-end success estimate, with an interval covering zero (Appendix~\ref{app:e2e}). Our measurements carry two-sided flips under compression \citep{dutta2024accuracy,jang2026flat} from whole episodes to single decisions and separate the action from the continuation that a per-step model swap changes together \citep{gonuguntla2026replay}.

\paragraph{Next measurement.} The telecom holdout leaves its cause open because domain and tasks changed together. Running all three holdout configurations on the 80 telecom development tasks over 20 blocks would separate a telecom domain effect from the particular holdout tasks. %

\clearpage %
\section*{Limitations}

Our measurements are offline values of one intervention per task and block, taken under simulated quantization on a deterministic serving stack, so deployed low-bit kernels, latency and the cost of a fallback lie outside their scope. The crossed forks cover Qwen3-4B and Qwen3-8B on \tautwo{}, with BFCL, ALFWorld and Mistral-Small-3.2-24B as scope checks (Appendix~\ref{app:scope}), and effects may differ for other model families and scales (Appendix~\ref{app:coverage}). All \tautwo{} episodes use one greedy user simulator, and individual rescue and damage labels changed under a second simulator, so the averages describe this simulator (Appendix~\ref{app:ordinary}). The telecom holdout covers 34 tasks and changed domain and tasks together, which leaves the cause of its negative estimate open (Appendix~\ref{app:prereg}). The common-state intervals still allow a BF16 advantage of a few points (Appendix~\ref{app:resolution}), and the protocol amendments made after interim looks are listed in Appendix~\ref{app:timeline}.

\section*{Ethics Statement}

This work involves no crowdworkers, human annotation or personal data. Where a user is present (\tautwo{}), it is simulated by an LLM. Tasks are public benchmark tasks, and all agents are openly released Qwen3, Mistral-Small-3.2 and (gate check only) Llama-3.1 checkpoints, or simulated-quantized versions of them. Outcomes use the automatic task-success endpoints defined in \S\ref{sec:setting}, including the database-only retail and airline checks, and action categories are assigned by fixed rules. Reported values map to committed analysis outputs (Appendix~\ref{app:figsrc}). %
The main risk we see is over-reading. A small forked-state gain on one simulator and a fixed task pool falls short of certifying a quantized agent for deployment, and our end-to-end comparisons fell short of equivalence. Deployment decisions in consequential settings need evidence beyond these measurements.

\appendix
\renewcommand{\dbltopfraction}{0.9}
\renewcommand{\dblfloatpagefraction}{0.55}
\setcounter{dbltopnumber}{3}
\makeatletter
\setlength{\@dblfptop}{0pt}
\setlength{\@dblfpsep}{16pt}
\setlength{\@dblfpbot}{0pt plus 1fil}
\makeatother
\clearpage
\section{Experimental Details}
\label{app:setup}

This appendix gives the implementation and evaluation details behind \S\ref{sec:method} and \S\ref{sec:setting}, the notation used in the later appendices, end-to-end scores, coverage and quantization configurations.
\label{app:figsrc}Every reported value maps to a committed analysis output or frozen manifest.

\subsection{Implementation and evaluation}

\paragraph{Splits.} The development split has 58 retail, 27 airline and 80 telecom tasks. The holdout has 34 telecom tasks, and its pre-registration hash is recorded in the split file. %
The telecom split was fixed from \tautwo{}'s 114-task base split by a stratified, salted hash before any run. %
The external benchmarks are BFCL multi-turn (base, miss\_func, miss\_param and long\_context, 197 items each) and ALFWorld (134 out-of-distribution games). Items were selected by hash in manifests committed before any model run. %

\paragraph{Success endpoints.} Retail and airline success is a final-database match with normal termination. \tautwo{}'s official reward additionally scores natural-language and communicate checks, which this endpoint omits. Telecom uses the official deterministic task-basis success. Context overflow and unparseable output count as failures, with the amended missing-data rules of the Mistral panel as the one exception (Appendix~\ref{app:mistral}). Infrastructure outages count as missing data. The primary analysis excludes incomplete blocks, and panel-specific exceptions are stated with their estimates.

\paragraph{Fork mechanics.} A fork agent returns the forced action for the one step that follows $h_t$. The real environment executes a tool call, and the live user simulator answers text, generating fresh observations instead of reusing those of the original trajectory. \tautwo{} resets step and error counters when it starts from a history. The fork restores them to their values at $t$, so the remaining budget equals the original. The recount reproduced every max\_steps and too\_many\_errors episode on the checked blocks. %

\paragraph{Replay acceptance.} In the engineering gate, one BB field, the official reward, was missing from older trajectories and was recomputed from the stored reward information before the gate was read. The QQ no-op, which reruns the quantized action with the quantized model continuing, matched the original outcome at all 924 sampled development states. Ordinary-state checks matched 328/328 sampled states and a separate 135/135 gate set. On the holdout, the initial 135-state sample and the later full replay of all 1202 divergent states matched (Appendix~\ref{app:prereg}). Prefix-cache state is a known source of trajectory divergence \citep{patodiya2026same}. The BF16 and quantized models run in separate processes, so neither model reuses the other's cache entries, and the replay checks above found no outcome change from cache state at any audited state. %
Acceptance counts for the same-precision reference are in Appendix~\ref{app:sameprec}.

\paragraph{Serving determinism.} Pairs of closed-loop runs of the same configuration on two nodes at different times reproduced whole episodes token for token on dense BF16 paths at tensor parallelism 1 and 2 with 1 to 16 concurrent requests (e.g.\ 257/257 and 1520/1520 episodes). Nine arms that two studies ran independently at the same states were byte-identical. Exactness under KV-cache preemption and on other models, GPUs or vLLM versions remains untested. A deployed AWQ kernel path with fp16 dequantization was not replay-exact, and all reported outcomes use the dense BF16 path instead. %

\paragraph{User simulator.} The serving manifests record Qwen3-30B-A3B-Instruct as the user simulator for every primary episode. %

\paragraph{Quantization configurations.} Table~\ref{tab:qconfig} (Appendix~\ref{app:qconfig}) lists the settings of every checkpoint.

\subsection{Notation}
\label{app:names}

\paragraph{Configurations, blocks and branches.} A \emph{configuration} is a model with a quantization setting. A \emph{block} is one replicate over the task pool with its own session-identifier line (nonce) in the system prompt, and a \emph{unit} is one task in one block. Prompt blocks are a researcher-chosen perturbation, separate from any deployment distribution. In formulas, B denotes the BF16 model, Q its quantized version and Q$'$ a second checkpoint at the same bit width. Branches are written by action source and continuing model (BB, BQ, QB, QQ). KQ is the key-order branch (Appendix~\ref{app:keyorder}). %

\paragraph{States.} A \emph{first divergent state} is, for each task, block and configuration, the first decision at which $a_{\mathrm{B}}$ and $a_{\mathrm{Q}}$ differ in kind (tool call, text, stop), tool name, call count or order, or normalised arguments. Differences in text only are skipped. An \emph{ordinary state} is one decision per task, drawn by a hash fixed before generation, in three blocks (six in the extended Qwen3-8B common-state design, Appendix~\ref{app:commonstate}). Ordinary states come from the quantized model's own trajectories, from BF16 trajectories shared by all configurations of a model, or from quantized trajectories shared in the same way. A \emph{common state} is an ordinary state of the common-state design, at which the BF16 model, the quantized model and the second checkpoint each supply the current action. That design was fixed at three blocks and extended to six by amendment. %

\paragraph{Action swap, rescues and damages.} The paired effect of the action swap under each continuation is $\DQ=\YY{BQ}-\YY{QQ}$ ($=G$) and $\DB=\YY{BB}-\YY{QB}$, each in $\{-1,0,+1\}$. When $\DQ\neq\DB$ (``off-diagonal''), the swap's value depends on the continuation, and the off-diagonal share is the fraction of such states. For the action swap, a \emph{rescue} is $(\YY{BQ},\YY{QQ})=(1,0)$ and a \emph{damage} is $(0,1)$. With rates $r$ and $d$, the turnover is $M=r+d$ and the net gain on the full pool is $\Gall=r-d$. $I=\DQ-\DB$ is the interaction of action and continuation (Table~\ref{tab:ghf}). $D_{cc}=\YY{QB}-\YY{BB}$ compares two quantization methods on the same ordinary states (Appendix~\ref{app:mistral}). $K(c)$ is the value of the BF16 action beyond the second checkpoint's action at a common state under continuation $c$ (Appendix~\ref{app:commonstate}). $M'$, $r'$, $d'$ and the off-diagonal share for the same-precision reference are defined likewise with the second checkpoint in place of the BF16 model. All contrasts measure effects on task success, and the correctness of an action is a separate question. %

\paragraph{Denominators and inference.} Three denominators are kept separate. (1)~\emph{Full pool}: every task$\times$block unit, with units lacking a first divergent state counted as zero. $r$, $d$, $M$ and $\Gall$ use it. (2)~\emph{Divergent states}: the first divergent states of a configuration. $G$, $H$, $F$ and the off-diagonal share use it. (3)~\emph{Ordinary states}: one state per task. Within a configuration all branches share each state and all tasks of a block share its nonce, so within-configuration contrasts are paired and identified. Cross-configuration comparisons are descriptive, because each configuration selects its own divergences. With $n_{\mathrm{blk}}$ complete blocks (only complete blocks enter), the primary \emph{fixed-pool} interval treats the task pool as fixed (block bootstrap if $n_{\mathrm{blk}}\ge10$, else block-$t$, a $t$ interval over block means). The \emph{new-task} interval is a crossed task$\times$block bootstrap (``crossed''). A contrast is \emph{resolved} under an interval when that interval excludes zero, and the interval is named. A contrast is \emph{unresolved} when the reported interval or intervals include zero. Where a frozen protocol sets its own reading rule, the rule is stated with the result. %

\subsection{End-to-end success}
\label{app:e2e}

The full-pool net gain and the gains of action replacement, takeover and joint replacement are local effects of one intervention, and end-to-end success is a separate comparison. Table~\ref{tab:e2e} gives it on the same checkpoints, the same tasks and the same 20 prompt blocks as the forks. %

\begin{table*}[t]
\centering%
\renewcommand{\arraystretch}{1.12}\fontsize{8.5}{10}\selectfont\setlength{\tabcolsep}{4pt}
\setlength{\tabcolsep}{4pt}
\begin{tabular}{@{}lrrcc@{}}
\toprule
Configuration & Q / BF16 (\%) & Q$-$BF16 [95\% CI] & NI 3\pp{} & TOST $\pm$3\pp{} \\
\midrule
4B GPTQ-W4 & 40.0 / 35.8 & $+4.2$ [$-2.4$, $+10.5$] & pass & fail \\
4B AWQ-style W4  & 36.8 / 35.8 & $+0.9$ [$-5.8$, $+7.5$] & fail & fail \\
8B GPTQ-W4 & 30.1 / 30.4 & $-0.2$ [$-7.7$, $+6.8$] & fail & fail \\
8B AWQ-style W4  & 31.5 / 30.4 & $+1.2$ [$-5.3$, $+7.6$] & fail & fail \\
pool, all four & & $+1.5$ [$-2.7$, $+5.6$] & pass & fail \\
\bottomrule
\end{tabular}
\caption{End-to-end closed-loop success of Qwen3-4B (4B) and Qwen3-8B (8B) configurations. Q / BF16: success rates of the quantized and BF16 models. Q$-$BF16: difference in pp with a crossed task$\times$block bootstrap 95\% confidence interval (CI), units paired on prompt block and the bootstrap drawn jointly across configurations. NI: non-inferiority of the quantized model at a 3\pp{} margin. TOST: equivalence by two one-sided tests at $\pm$3\pp{}.}
\label{tab:e2e}
\end{table*}

\subsection{Coverage}
\label{app:coverage}

Table~\ref{tab:coverage} lists the tasks, states and complete blocks behind the Qwen3 results. Qwen3-14B and Qwen3-32B also have end-to-end runs, and the Qwen3-14B row is the only crossed configuration at that scale. These runs stay outside the reported analyses. A Llama-3.1-8B-Instruct check on BFCL (GPTQ-W4) stopped at its gate because BF16 solved 5 of 120 gate items, so its four branches carried almost no outcome variance. %
 
\begin{table*}[t]
\centering%
\renewcommand{\arraystretch}{1.12}\fontsize{8.5}{10}\selectfont\setlength{\tabcolsep}{4pt}
\begin{tabular}{@{}P{2.4cm}P{2.25cm}P{2.25cm}P{2.1cm}P{2.3cm}P{2.65cm}@{}}
\toprule
Configuration & \tautwo{} retail (first div.) & \tautwo{} airline (first div.) & telecom dev (first div.) & Ordinary (retail / airline) & BFCL / ALFWorld \\
\midrule
4B GPTQ-W4 & \shortstack[l]{58 t / 727 st\\{}20/20 blocks} & \shortstack[l]{27 t / 375 st\\{}20/20 blocks} & holdout only & \shortstack[l]{3 blocks\\{}393 / 177 rows} & 788 items $\cdot$ 573 st / -- \\
4B AWQ-style W4  & \shortstack[l]{58 t / 986 st\\{}20/20 blocks} & \shortstack[l]{27 t / 453 st\\{}20/20 blocks} & -- & \shortstack[l]{3 blocks\\{}393 / 177 rows} & 788 items $\cdot$ 642 st / -- \\
4B KV8     & \shortstack[l]{58 t / 412 st\\{}22/22 blocks} & \shortstack[l]{27 t / 181 st\\{}22/22 blocks} & -- & \shortstack[l]{3 blocks\\{}393 / 177 rows} & -- / -- \\
8B GPTQ-W4 & \shortstack[l]{58 t / 1017 st\\{}20/20 blocks} & \shortstack[l]{27 t / 456 st\\{}20/20 blocks} & \shortstack[l]{80 t / 477 st\\{}6/6 blocks}; holdout & \shortstack[l]{3 blocks\\{}393 / 177 rows} & 788 $\cdot$ 533 st / 134 $\cdot$ 115 st \\
8B AWQ-style W4  & \shortstack[l]{58 t / 1087 st\\{}20/20 blocks} & \shortstack[l]{27 t / 475 st\\{}20/20 blocks} & \shortstack[l]{80 t / 480 st\\{}6/6 blocks}; holdout & \shortstack[l]{3 blocks\\{}393 / 177 rows} & 788 $\cdot$ 548 st / 134 $\cdot$ 119 st \\
8B KV8     & \shortstack[l]{58 t / 585 st\\{}22/22 blocks} & \shortstack[l]{27 t / 270 st\\{}22/22 blocks} & -- & \shortstack[l]{3 blocks\\{}393 / 177 rows} & -- / -- \\ %
\midrule
4B GPTQ-W3 & \shortstack[l]{58 t / 162 st\\{}3/3 blocks} & \shortstack[l]{27 t / 70 st\\{}3/3 blocks} & -- & -- & 591 $\cdot$ 540 st$^\dagger$ / -- \\ %
4B KV4-rot & \shortstack[l]{58 t / 164 st\\{}3/3 blocks} & \shortstack[l]{27 t / 69 st\\{}3/3 blocks} & -- & -- & -- / -- \\
8B GPTQ-W3 & \shortstack[l]{58 t / 166 st\\{}3/3 blocks} & \shortstack[l]{27 t / 76 st\\{}3/3 blocks} & -- & -- & 788 $\cdot$ 696 st / 134 $\cdot$ 130 st \\
8B KV4-rot & \shortstack[l]{58 t / 165 st\\{}3/3 blocks} & \shortstack[l]{27 t / 74 st\\{}3/3 blocks} & -- & -- & -- / -- \\
8B RTN-W3  & \shortstack[l]{58 t / 174 st\\{}3/3 blocks} & \shortstack[l]{27 t / 81 st\\{}3/3 blocks} & -- & -- & -- / -- \\
4B / 8B KV4 (unrotated) & -- & -- & -- & -- & 788 $\cdot$ 787 / 788 $\cdot$ 788 st; 8B ALFWorld 134 $\cdot$ 134 st \\
\midrule
14B GPTQ-W4 & -- & -- & -- & -- & 788 $\cdot$ 471 st / -- %
 \\
\bottomrule
\end{tabular}
\caption{Completed coverage of the crossed interventions for Qwen3-4B (4B), Qwen3-8B (8B) and Qwen3-14B (14B), counted after the strict admission rule. First div.: first divergent states. t = tasks, st = states, blocks = complete/found blocks. Ordinary: blocks completed and state rows per domain, with the first three blocks analysed. BFCL and ALFWorld: items $\cdot$ states, one run each. ``holdout'': the configuration is in the telecom holdout (Appendix~\ref{app:prereg}), and the telecom development results (Appendix~\ref{app:holdoutdev}) use the six blocks listed. Settings below the first rule are stress settings. $^\dagger$The QQ no-op rule (Appendix~\ref{app:setup}) dropped the BFCL base category for 4B GPTQ-W3.}
\label{tab:coverage}
\end{table*}

\subsection{Quantization configurations}
\label{app:qconfig}

Table~\ref{tab:qconfig} gives the checkpoint settings from metadata and quantization code. The metadata omit the AWQ-style row-sample size (4096 is the code default) and the KV-cache scale dtype (the hook computes in fp32). Mistral has a single calibration draw. %

\begin{table*}[t]
\centering%
\renewcommand{\arraystretch}{1.12}
\fontsize{8.5}{10}\selectfont
\setlength{\tabcolsep}{4pt}
\begin{tabular}{@{}P{2.15cm}P{2.15cm}P{1.55cm}P{1.0cm}P{4.0cm}P{3.7cm}@{}}
\toprule
Config.\ (models) & Method, weights & Group, grid & Act-order & Calibration & KV cache / activations / rotation \\
\midrule
GPTQ-W4 Q (4B, 8B) & GPTQ, W4; damp 0.01, block 128, true-sequential & 128, asym. & yes & draw c1: 57 BF16 agent-trajectory states (39 retail, 18 airline calibration tasks, disjoint from the 85 dev tasks), one state per task, full context (median 6041 / 6310 tok.); 64{,}000 activation rows (1122--1123 per state); draw seed 20260729 & KV BF16, act.\ BF16, no rotation \\
GPTQ-W4 Q$'$ (4B, 8B) & as Q & 128, asym. & yes & draw c2, seed 20260730: same 57 tasks and trajectories, re-drawn decision state and rows (5/57 states shared with c1); not disjoint & as Q \\
AWQ-style W4 (4B, 8B) & AWQ-style: per-linear scale $\bar{|x|}^\alpha$, 20-point $\alpha$ grid, scale folded into weight, then RTN; no clip search & 128, asym. & n/a (GPTQ only) & draw c1 as GPTQ-W4 Q; $\alpha$ fit on a 4096-row sample (code default; per-run value not recorded) & as Q \\
GPTQ-W3$^s$ Q / Q$'$ (4B, 8B) & GPTQ, W3 & 128, asym. & yes & c1 / c2 as W4 & as Q \\
RTN-W3$^s$ (4B, 8B) & RTN, W3 & 128, asym. & n/a & none & as Q \\
\addlinespace[3pt]
KV8 (4B, 8B) & BF16 weights & -- & -- & none & INT8 K and V, asym.\ min--max per (token, KV head) over head\_dim, no clip, no rotation; fake quant in fp32 \\
KV4-rot$^s$ (4B, 8B) & BF16 weights & -- & -- & none & INT4 K and V, same grid as KV8; online normalized Hadamard on head\_dim: q and k rotated, V rotated and rotated back \\
\addlinespace[3pt]
Mistral GPTQ-W4 / RTN-W4 (Mistral-Small-3.2-24B) & GPTQ W4 / RTN W4 & 128, asym. & yes / n/a & GPTQ: draw c1, 57 states, 64{,}000 rows, seed 20260729 (no c2 draw) & as Q \\
\bottomrule
\end{tabular}
\caption{Quantization configuration of every checkpoint, from the checkpoint metadata. 4B and 8B denote Qwen3-4B and Qwen3-8B. W$n$: $n$-bit weights, KV$n$: $n$-bit KV cache, INT8/INT4: 8-/4-bit integers, RTN: round-to-nearest. AWQ-style W4 folds the scale into the weight, then applies RTN without clip search. Q: the quantized model. Q$'$: the second checkpoint, which re-draws the decision state and activation rows from the same 57 calibration trajectories. One Qwen3-8B calibration context in draw c1 is marked truncated. All checkpoints are fake-quantized and stored in BF16, with embeddings, \texttt{lm\_head} and norms kept in BF16. $^s$Stress configuration.}
\label{tab:qconfig}
\end{table*}

\subsection{Protocol fixing}
\label{app:timeline}

Each experiment fixed its state restoration, branching, configurations, blocks and acceptance rules before its outcomes were read, and analyses chosen later are marked post hoc where they appear. Fixing an analysis before new outcomes are read leaves in place what earlier outcomes had already shown, and no amendment below adds a new confirmation of the W4 benefit estimates. The experiments were designed in sequence, so later designs were informed by the outcomes of earlier ones, and the main consequence and pattern summaries of the development forks are exploratory. The telecom holdout fixed its tasks, configurations, weights, blocks, estimator and readings before any run, and its full QQ replay was a later audit. Amendments are dated and reported with their results. After an interim look, the same-precision reference replaced its ``similar'' reading by stricter equivalence tests, with estimates unchanged (Appendix~\ref{app:sameprec}). The common-state design was extended from three to six blocks under a two-look rule fixed after an unresolved first look, when some state outcomes were already known (Appendix~\ref{app:commonstate}). The key-order reference froze eligibility and analysis before its outcomes were read and reused the original QQ and BF16 branch outcomes (Appendix~\ref{app:keyorder}). The frozen margin rule was selected on one development block and validated on later blocks of the same tasks (Appendix~\ref{app:r3}). The Mistral failure-scoring rule was added after the first phase of that panel (Appendix~\ref{app:mistral}).
\section{Additional Results for Tasks Saved and Tasks Spoiled}
\label{app:tables}

This appendix gives the per-configuration values behind \S\ref{sec:value}. Branches are named by action source and continuing model, with B for the BF16 model and Q for the quantized model (BB, BQ, QB, QQ), and $r$, $d$ and $M=r+d$ denote the rescue, damage and turnover rates. Table rows label Qwen3-4B and Qwen3-8B as 4B and 8B.

\subsection{Rescues and damages per configuration}

Table~\ref{tab:r1} expands Figure~\ref{fig:rd}a with every W4 and KV8 configuration and the five stress configurations, and reports both interval types. The full-pool net gain ($\Gall$) counts every task--block unit, so units without a divergence enter as zero. It measures the local effect of replacing one action and differs from the end-to-end success difference between the two models (Appendix~\ref{app:e2e}).

\begin{table*}[t]
\centering%
\renewcommand{\arraystretch}{1.12}\fontsize{8.5}{10}\selectfont\setlength{\tabcolsep}{4pt}
\begin{tabular}{@{}lrrrrrll@{}}
\toprule
Configuration & units & div. & $r$ & $d$ & $M$ & $\Gall$ [fixed pool] & [new tasks] \\
\midrule
4B GPTQ-W4 & 85$\times$20 & 1102 & 6.8 & 8.7 & 15.5 & $-1.9$ [$-3.7$, $-0.1$] & [$-6.8$, $+2.7$] \\
4B AWQ-style W4  & 85$\times$20 & 1439 & 9.8 & 10.1 & 19.9 & $-0.3$ [$-2.2$, $+1.6$] & [$-5.4$, $+5.2$] \\
8B GPTQ-W4 & 85$\times$20 & 1473 & 10.4 & 7.5 & 17.9 & $+2.9$ [$+1.4$, $+4.4$] & [$-2.9$, $+9.2$] \\
8B AWQ-style W4  & 85$\times$20 & 1562 & 11.4 & 9.5 & 20.9 & $+1.9$ [$-0.6$, $+4.3$] & [$-3.9$, $+8.1$] \\
W4 pool    & 85$\times$20 & & 9.6 & 9.0 & 18.6 & $+0.7$ [$-0.3$, $+1.6$] & [$-2.1$, $+3.4$] \\
\midrule
4B KV8 & 85$\times$22 & 593 & 3.1 & 3.4 & 6.5 & $-0.3$ [$-1.7$, $+1.0$] & [$-3.0$, $+2.3$] \\
8B KV8 & 85$\times$22 & 855 & 4.3 & 3.7 & 8.1 & $+0.6$ [$-0.4$, $+1.6$] & [$-1.9$, $+3.0$] \\
\midrule
4B GPTQ-W3$^s$ & 85$\times$3 & 232 & 9.0 & 10.2 & 19.2 & $-1.2$ [$-7.0$, $+4.7$] & [$-8.6$, $+5.9$] \\
4B KV4-rot$^s$ & 85$\times$3 & 233 & 11.8 & 7.1 & 18.8 & $+4.7$ [$-4.1$, $+13.5$] & [$-3.1$, $+12.9$] \\
8B GPTQ-W3$^s$ & 85$\times$3 & 242 & 10.2 & 8.2 & 18.4 & $+2.0$ [$-12.5$, $+16.4$] & [$-6.3$, $+11.8$] \\
8B KV4-rot$^s$ & 85$\times$3 & 239 & 11.8 & 8.6 & 20.4 & $+3.1$ [$-14.5$, $+20.8$] & [$-5.9$, $+12.9$] \\
8B RTN-W3$^s$ & 85$\times$3 & 255 & 3.1 & 3.5 & 6.7 & $-0.4$ [$-3.8$, $+3.0$] & [$-4.7$, $+3.9$] \\
\bottomrule
\end{tabular}
\caption{Action-only replacement at the quantized model's first divergent state with the quantized model continuing, \tautwo{} retail and airline development panel (pp of units). Units: tasks $\times$ complete blocks. Div.: number of first divergent states. $r$, $d$, $M$: rescue, damage and turnover rates over all units, with units lacking a divergent state counted as zero. $\Gall=r-d$. Fixed-pool 95\% intervals use a block bootstrap ($\ge$10 blocks) or a $t$ interval over block means. New-task 95\% intervals use a crossed task$\times$block bootstrap. W3: 3-bit weights. KV8: INT8 KV cache. KV4-rot: rotated 4-bit KV cache. RTN: round-to-nearest. $^s$Stress configuration, three blocks.}
\label{tab:r1}
\end{table*}

\subsection{Three upgrade modes}
\label{app:r2main}

Table~\ref{tab:ghf} reports action replacement ($G$), takeover ($H$), joint replacement ($F$) and the difference between action replacement and takeover ($G-H$) on the first divergent states of each configuration, the values behind Figure~\ref{fig:rd}c and Table~\ref{tab:scope}. Table~\ref{tab:ghf-patterns} adds the contrasts $F-G$ and $F-H$ and the share of states at which the value of the BF16 action depends on which model continues, which is how often a rescue under one mode fails under another. The four W4 tests of $G-H$ form one Holm family, in which 4B GPTQ-W4 has adjusted $p=.069$. %

\begin{table*}[t]
\centering\fontsize{9}{11}\selectfont
\renewcommand{\arraystretch}{1.15}\setlength{\tabcolsep}{4pt} %
\begin{tabular*}{\textwidth}{@{\extracolsep{\fill}}llllll@{}}
\toprule
Configuration & states & $G$ (action) & $H$ (continuation) & $F$ (joint) & $G-H$ \\
\midrule
4B GPTQ-W4 & 1102 & \shortstack[l]{$-2.9$ [$-5.7$, $-0.2$]\\{}[$-10.3$, $+4.2$]} & \shortstack[l]{$+0.7$ [$-1.6$, $+3.0$]\\{}[$-4.0$, $+6.0$]} & \shortstack[l]{$-2.5$ [$-5.3$, $+0.2$]\\{}[$-9.9$, $+4.7$]} & \shortstack[l]{$-3.6$ [$-6.6$, $-0.6$]\\{}[$-11.2$, $+3.6$]} \\
\addlinespace[1pt] %
4B AWQ-style W4 & 1439 & \shortstack[l]{$-0.3$ [$-2.6$, $+1.9$]\\{}[$-6.5$, $+6.1$]} & \shortstack[l]{$+0.9$ [$-1.1$, $+2.9$]\\{}[$-3.7$, $+5.5$]} & \shortstack[l]{$-0.8$ [$-3.3$, $+1.6$]\\{}[$-7.4$, $+5.8$]} & \shortstack[l]{$-1.3$ [$-3.0$, $+0.6$]\\{}[$-7.7$, $+5.1$]} \\
\addlinespace[1pt] %
8B GPTQ-W4 & 1473 & \shortstack[l]{$+3.3$ [$+1.6$, $+5.0$]\\{}[$-3.3$, $+10.5$]} & \shortstack[l]{$+0.4$ [$-1.2$, $+1.9$]\\{}[$-3.8$, $+4.9$]} & \shortstack[l]{$+5.4$ [$+3.5$, $+7.2$]\\{}[$-2.0$, $+13.0$]} & \shortstack[l]{$+2.9$ [$+1.3$, $+4.7$]\\{}[$-3.2$, $+9.3$]} \\
\addlinespace[1pt] %
8B AWQ-style W4 & 1562 & \shortstack[l]{$+2.1$ [$-0.6$, $+4.7$]\\{}[$-4.1$, $+8.7$]} & \shortstack[l]{$-1.3$ [$-3.5$, $+0.9$]\\{}[$-7.1$, $+4.6$]} & \shortstack[l]{$+0.6$ [$-1.5$, $+2.7$]\\{}[$-6.2$, $+7.3$]} & \shortstack[l]{$+3.4$ [$+1.5$, $+5.2$]\\{}[$-1.6$, $+8.6$]} \\
\addlinespace[1pt] %
W4 pool & 5576 & \shortstack[l]{$+0.5$ [$-0.7$, $+1.7$]\\{}[$-2.9$, $+3.8$]} & \shortstack[l]{$+0.2$ [$-1.3$, $+1.6$]\\{}[$-2.7$, $+2.8$]} & \shortstack[l]{$+0.7$ [$-0.6$, $+2.1$]\\{}[$-3.1$, $+4.4$]} & \shortstack[l]{$+0.4$ [$-0.8$, $+1.7$]\\{}[$-3.1$, $+3.7$]} \\
\addlinespace[1pt] %
\midrule
4B KV8 & 593 & \shortstack[l]{$-1.0$ [$-5.3$, $+3.2$]\\{}[$-9.4$, $+7.3$]} & \shortstack[l]{$-2.5$ [$-4.5$, $-0.5$]\\{}[$-6.4$, $+1.1$]} & \shortstack[l]{$-1.2$ [$-4.7$, $+2.3$]\\{}[$-9.3$, $+7.7$]} & \shortstack[l]{$+1.5$ [$-3.1$, $+6.0$]\\{}[$-7.1$, $+9.7$]} \\
\addlinespace[1pt] %
8B KV8 & 855 & \shortstack[l]{$+1.3$ [$-0.9$, $+3.4$]\\{}[$-4.0$, $+6.9$]} & \shortstack[l]{$-0.4$ [$-2.1$, $+1.5$]\\{}[$-3.5$, $+2.9$]} & \shortstack[l]{$+2.5$ [$+0.0$, $+5.1$]\\{}[$-3.5$, $+8.1$]} & \shortstack[l]{$+1.6$ [$-0.6$, $+3.9$]\\{}[$-3.8$, $+7.2$]} \\
\addlinespace[1pt] %
KV8 pool & 1448 & \shortstack[l]{$+0.1$ [$-2.8$, $+3.0$]\\{}[$-5.3$, $+5.6$]} & \shortstack[l]{$-1.4$ [$-2.7$, $-0.1$]\\{}[$-3.7$, $+0.9$]} & \shortstack[l]{$+0.6$ [$-1.8$, $+3.1$]\\{}[$-4.5$, $+5.8$]} & \shortstack[l]{$+1.6$ [$-1.2$, $+4.1$]\\{}[$-3.8$, $+6.6$]} \\
\addlinespace[1pt] %
\midrule
4B GPTQ-W3$^s$ & 232 & \shortstack[l]{$-1.3$ [$-7.7$, $+5.0$]\\{}[$-9.7$, $+6.6$]} & \shortstack[l]{$+4.3$ [$-10.7$, $+19.4$]\\{}[$-6.6$, $+14.5$]} & \shortstack[l]{$+1.3$ [$-1.9$, $+4.4$]\\{}[$-6.8$, $+9.4$]} & \shortstack[l]{$-5.6$ [$-15.0$, $+3.7$]\\{}[$-15.0$, $+4.0$]} \\
\addlinespace[1pt] %
4B KV4-rot$^s$ & 233 & \shortstack[l]{$+5.2$ [$-4.5$, $+14.9$]\\{}[$-3.4$, $+14.2$]} & \shortstack[l]{$+9.4$ [$+5.1$, $+13.7$]\\{}[$+2.1$, $+17.0$]} & \shortstack[l]{$+12.4$ [$+11.0$, $+13.9$]\\{}[$+3.0$, $+21.6$]} & \shortstack[l]{$-4.3$ [$-18.3$, $+9.8$]\\{}[$-13.5$, $+7.0$]} \\
\addlinespace[1pt] %
8B GPTQ-W3$^s$ & 242 & \shortstack[l]{$+2.1$ [$-13.3$, $+17.2$]\\{}[$-6.7$, $+12.1$]} & \shortstack[l]{$+0.4$ [$-7.4$, $+8.2$]\\{}[$-7.2$, $+8.1$]} & \shortstack[l]{$+2.1$ [$-7.4$, $+11.4$]\\{}[$-6.6$, $+11.8$]} & \shortstack[l]{$+1.7$ [$-9.1$, $+12.3$]\\{}[$-7.7$, $+11.2$]} \\
\addlinespace[1pt] %
8B KV4-rot$^s$ & 239 & \shortstack[l]{$+3.3$ [$-15.4$, $+21.9$]\\{}[$-6.6$, $+13.8$]} & \shortstack[l]{$-0.8$ [$-18.2$, $+16.3$]\\{}[$-11.2$, $+8.5$]} & \shortstack[l]{$+4.6$ [$-16.0$, $+25.0$]\\{}[$-6.6$, $+15.9$]} & \shortstack[l]{$+4.2$ [$+2.5$, $+5.8$]\\{}[$-3.3$, $+12.2$]} \\
\addlinespace[1pt] %
8B RTN-W3$^s$ & 255 & \shortstack[l]{$-0.4$ [$-3.8$, $+3.0$]\\{}[$-4.7$, $+3.9$]} & \shortstack[l]{$+15.3$ [$+0.7$, $+29.9$]\\{}[$+5.9$, $+25.9$]} & \shortstack[l]{$+18.4$ [$+16.7$, $+20.1$]\\{}[$+9.4$, $+28.2$]} & \shortstack[l]{$-15.7$ [$-27.5$, $-3.9$]\\{}[$-24.7$, $-7.1$]} \\
\addlinespace[1pt] %
\bottomrule
\end{tabular*}
\caption{Net gain of the three upgrade modes on first divergent states (pp). $G$, $H$ and $F$ compare action-only (BQ), continuation-only (QB) and joint (BB) replacement with QQ, and $G-H$ compares the first two. In each two-line cell, the first line gives the estimate and its fixed-pool 95\% interval and the second line the new-task 95\% interval (crossed task$\times$block bootstrap). Fixed-pool intervals use a block bootstrap over 20--22 blocks, or a $t$ interval over three blocks for the stress configurations ($^s$). Pools average the configuration estimates with equal weight, and stress configurations are reported separately. 4B and 8B denote Qwen3-4B and Qwen3-8B.}
\label{tab:ghf}
\label{tab:ghf-crossed}
\end{table*}

\begin{table*}[t]
\centering\fontsize{9}{11}\selectfont
\renewcommand{\arraystretch}{1.15}\setlength{\tabcolsep}{4pt} %
\begin{tabular*}{\textwidth}{@{\extracolsep{\fill}}llllll@{}}
\toprule
Configuration & $F-G$ & $F-H$ & off-diag.\ \% & rev.\ \% & mean $I$ \\
\midrule
4B GPTQ-W4 & \shortstack[l]{$+0.5$ [$-1.2$, $+2.2$]\\{}[$-4.7$, $+6.1$]} & \shortstack[l]{$-3.2$ [$-6.4$, $+0.2$]\\{}[$-11.1$, $+4.4$]} & 25.0 & 2.6 & $+0.3$ \\
\addlinespace[1pt] %
4B AWQ-style W4 & \shortstack[l]{$-0.4$ [$-2.4$, $+1.5$]\\{}[$-6.2$, $+5.4$]} & \shortstack[l]{$-1.7$ [$-3.6$, $+0.4$]\\{}[$-8.7$, $+4.7$]} & 29.3 & 3.1 & $+1.3$ \\
\addlinespace[1pt] %
8B GPTQ-W4 & \shortstack[l]{$+2.1$ [$+0.2$, $+4.0$]\\{}[$-3.0$, $+7.1$]} & \shortstack[l]{$+5.0$ [$+3.2$, $+6.8$]\\{}[$-2.0$, $+12.5$]} & 22.9 & 2.7 & $-1.7$ \\
\addlinespace[1pt] %
8B AWQ-style W4 & \shortstack[l]{$-1.5$ [$-4.2$, $+0.8$]\\{}[$-7.6$, $+4.2$]} & \shortstack[l]{$+1.9$ [$-0.1$, $+3.8$]\\{}[$-3.8$, $+7.5$]} & 30.5 & 4.2 & $+0.3$ \\
\addlinespace[1pt] %
W4 pool & \shortstack[l]{$+0.2$ [$-1.0$, $+1.2$]\\{}[$-2.8$, $+3.1$]} & \shortstack[l]{$+0.5$ [$-1.0$, $+2.1$]\\{}[$-3.5$, $+4.5$]} & 26.9 & 3.2 & $+0.0$ \\
\addlinespace[1pt] %
\midrule
4B KV8 & \shortstack[l]{$-0.2$ [$-3.5$, $+3.7$]\\{}[$-5.7$, $+5.3$]} & \shortstack[l]{$+1.3$ [$-2.5$, $+4.9$]\\{}[$-6.7$, $+9.5$]} & 15.7 & 1.2 & $-2.4$ \\
\addlinespace[1pt] %
8B KV8 & \shortstack[l]{$+1.2$ [$-0.5$, $+3.2$]\\{}[$-2.5$, $+5.0$]} & \shortstack[l]{$+2.8$ [$+0.3$, $+5.5$]\\{}[$-3.1$, $+8.7$]} & 14.9 & 1.2 & $-1.5$ \\
\addlinespace[1pt] %
KV8 pool & \shortstack[l]{$+0.5$ [$-1.6$, $+2.8$]\\{}[$-2.8$, $+3.9$]} & \shortstack[l]{$+2.1$ [$-0.2$, $+4.2$]\\{}[$-3.0$, $+7.3$]} & 15.3 & 1.2 & $-1.9$ \\
\addlinespace[1pt] %
\midrule
4B GPTQ-W3$^s$ & \shortstack[l]{$+2.6$ [$-0.6$, $+5.8$]\\{}[$-5.2$, $+10.5$]} & \shortstack[l]{$-3.0$ [$-15.1$, $+9.1$]\\{}[$-12.2$, $+6.6$]} & 31.5 & 3.9 & $+1.7$ \\
\addlinespace[1pt] %
4B KV4-rot$^s$ & \shortstack[l]{$+7.3$ [$-2.0$, $+16.5$]\\{}[$-1.3$, $+17.1$]} & \shortstack[l]{$+3.0$ [$-1.9$, $+8.0$]\\{}[$-6.2$, $+12.2$]} & 33.0 & 1.7 & $+2.1$ \\
\addlinespace[1pt] %
8B GPTQ-W3$^s$ & \shortstack[l]{$+0.0$ [$-11.0$, $+11.1$]\\{}[$-9.1$, $+9.1$]} & \shortstack[l]{$+1.7$ [$-0.1$, $+3.4$]\\{}[$-6.8$, $+11.0$]} & 23.1 & 4.5 & $+0.4$ \\
\addlinespace[1pt] %
8B KV4-rot$^s$ & \shortstack[l]{$+1.3$ [$-1.8$, $+4.3$]\\{}[$-7.2$, $+9.9$]} & \shortstack[l]{$+5.4$ [$+0.8$, $+10.0$]\\{}[$-2.5$, $+14.6$]} & 31.0 & 4.6 & $-2.1$ \\
\addlinespace[1pt] %
8B RTN-W3$^s$ & \shortstack[l]{$+18.8$ [$+13.8$, $+23.9$]\\{}[$+10.2$, $+27.8$]} & \shortstack[l]{$+3.1$ [$-13.0$, $+19.2$]\\{}[$-7.1$, $+13.3$]} & 29.0 & 0.4 & $-3.5$ \\
\addlinespace[1pt] %
\bottomrule
\end{tabular*}
\caption{Paired mode contrasts and outcome patterns on the states of Table~\ref{tab:ghf}. Two-line cells give the estimate with its fixed-pool 95\% interval on the first line and the new-task 95\% interval on the second, both unadjusted (pp). $\DQ=\YY{BQ}-\YY{QQ}$ and $\DB=\YY{BB}-\YY{QB}$ are the effects of swapping in the BF16 action with the quantized model and with the BF16 model continuing. Off-diag.: \% of states with $\DQ\ne\DB$. Rev.: \% of states with $\DQ=-\DB\ne0$. Mean $I$: mean of the interaction $I=\DQ-\DB$. $G-H$ compares the two upgrades with each other, and each upgrade's value against QQ appears in Table~\ref{tab:ghf}.}
\label{tab:ghf-patterns}
\end{table*}

\subsection{Joint rescue and rescue undone}
\label{app:offdiag}

The analyses in this subsection were chosen after the outcomes were seen. Each divergent state has four outcomes, written in the order BB, BQ, QB, QQ, and $k$ is the number of successful branches. A \emph{joint rescue} is the pattern \texttt{1000}, where only replacing both action and continuation succeeds. A \emph{rescue undone} is \texttt{0100} or \texttt{0110}, where the BF16 action succeeds with the quantized model continuing and fails with the BF16 model continuing.

\paragraph{Label-permutation reference.} The reference keeps $k$ for every state and permutes the four outcome labels uniformly. Under it, \texttt{1000} has probability $1/4$ at $k=1$, and rescue undone has probability $1/4$ at $k=1$ plus $1/6$ at $k=2$. The reference shares are $P(k{=}1)/4$ and $P(k{=}1)/4+P(k{=}2)/6$, computed on the same states and bootstrap draws as the observed shares (Table~\ref{tab:patternref}). %

\paragraph{Results.} In the \tautwo{} W4 pool, the observed share was 4.0\% for joint rescue (fixed pool [3.5, 4.5], new tasks [2.7, 5.3]) and 4.8\% for rescue undone (fixed pool [4.1, 5.4], new tasks [3.2, 6.5]). %
Joint rescue exceeded the reference by $+0.47$\pp{} with a fixed-pool interval above zero ([$+0.06$, $+0.87$]) and a new-task interval covering zero ([$-0.33$, $+1.37$]). Rescue undone fell below the reference ($-0.99$, fixed pool [$-1.58$, $-0.42$], new tasks [$-2.19$, $+0.27$]). The 8B configurations carried the joint-rescue excess. For rescue undone, every group's new-task interval covered zero or lay below it. The reference keeps $k$ and leaves execution unmodelled. A share near the reference means that the arrangement of successes is about as common as a random arrangement of the same number of successes, and it leaves open whether the individual outcomes are noise and whether the continuation matters. %

\begin{table*}[t]
\centering%
\renewcommand{\arraystretch}{1.12}\fontsize{8.5}{10}\selectfont\setlength{\tabcolsep}{4pt}
\begin{tabular*}{\textwidth}{@{\extracolsep{\fill}}llllll@{}}
\toprule
\multicolumn{6}{@{}l}{\textbf{A. Joint rescue}} \\
\addlinespace[2pt]
Group & states & obs. & ref. & obs.$-$ref.\ [fixed pool] & [new tasks] \\
\midrule
\tautwo{} W4 pool & 5576 & 3.96 & 3.48 & $+0.47$ [$+0.06$, $+0.87$] & [$-0.33$, $+1.37$] \\
\quad 4B W4 configurations & 2541 & 3.22 & 3.56 & $-0.34$ [$-0.83$, $+0.13$] & [$-1.46$, $+0.88$] \\
\quad 8B W4 configurations & 3035 & 4.69 & 3.40 & $+1.29$ [$+0.79$, $+1.78$] & [$-0.07$, $+2.74$] \\
\tautwo{} KV8 pool & 1448 & 2.99 & 2.17 & $+0.81$ [$+0.10$, $+1.52$] & [$-0.51$, $+2.19$] \\
Telecom 8B W4 pool & 957 & 3.24 & 3.94 & $-0.70$ [$-1.46$, $+0.05$] & [$-2.47$, $+1.15$] \\
Same-precision W4 (Q$'$ as B) & 784 & 2.27 & 3.50 & $-1.23$ [$-2.47$, $-0.04$] & [$-2.88$, $+0.51$] \\
BF16, same blocks & 784 & 4.96 & 3.53 & $+1.43$ [$-0.40$, $+3.42$] & [$-0.90$, $+4.26$] \\
Mistral, own BF16 states$^m$ & 464 & 3.4 & 2.9 & $+0.5$ [$-2.4$, $+3.5$] & [$-2.0$, $+3.6$] \\
\bottomrule
\end{tabular*}
\par\vspace{6pt}
\begin{tabular*}{\textwidth}{@{\extracolsep{\fill}}llllll@{}}
\toprule
\multicolumn{6}{@{}l}{\textbf{B. Rescue undone}} \\
\addlinespace[2pt]
Group & states & obs. & ref. & obs.$-$ref.\ [fixed pool] & [new tasks] \\
\midrule
\tautwo{} W4 pool & 5576 & 4.80 & 5.78 & $-0.99$ [$-1.58$, $-0.42$] & [$-2.19$, $+0.27$] \\
\quad 4B W4 configurations & 2541 & 4.77 & 6.01 & $-1.24$ [$-2.09$, $-0.45$] & [$-2.97$, $+0.69$] \\
\quad 8B W4 configurations & 3035 & 4.82 & 5.56 & $-0.74$ [$-1.35$, $-0.07$] & [$-2.34$, $+0.98$] \\
\tautwo{} KV8 pool & 1448 & 2.31 & 4.58 & $-2.28$ [$-2.98$, $-1.58$] & [$-3.62$, $-0.98$] \\
Telecom 8B W4 pool & 957 & 4.18 & 5.13 & $-0.95$ [$-1.70$, $-0.20$] & [$-2.63$, $+0.61$] \\
Same-precision W4 (Q$'$ as B) & 784 & 5.77 & 5.75 & $+0.02$ [$-2.33$, $+2.25$] & [$-2.66$, $+2.92$] \\
BF16, same blocks & 784 & 3.82 & 6.14 & $-2.32$ [$-4.46$, $-0.19$] & [$-4.68$, $+0.17$] \\
Mistral, own BF16 states$^m$ & 464 & 4.0 & 4.7 & $-0.7$ [$-2.0$, $+0.7$] & [$-3.0$, $+1.9$] \\
\bottomrule
\end{tabular*}
\par\vspace{6pt}
\caption{Joint rescue and rescue undone against the label-permutation reference that keeps the number $k$ of successful branches per state (\% of divergent states). Obs.: observed share. Ref.: reference share. Fixed-pool 95\% intervals use a block bootstrap for the 20- and 22-block \tautwo{} pools and a $t$ interval over per-block ratios for pools with fewer than 10 blocks (six, or five for Mistral). New-task 95\% intervals use a crossed task$\times$block bootstrap. The same-precision rows each fork at their own first divergence. $^m$States at which the Mistral actions differ on BF16 trajectories, so BB is the BF16 original, five blocks. These rows are kept separate from the first-divergent-state rows.}
\label{tab:patternref}
\end{table*}

\section{Telecom Holdout}
\label{app:prereg}

This appendix gives the registered design, the full results and the post hoc checks of the telecom holdout in \S\ref{sec:value}. 4B and 8B denote Qwen3-4B and Qwen3-8B, QQ is the baseline branch in which the quantized model keeps its own action and continues, and $r$, $d$ and $\Gall=r-d$ are the rescue rate, damage rate and full-pool net gain of action-only replacement.

\paragraph{Registered design.} The protocol was fixed before any holdout unit was run, and its hash was recorded in the split manifest. The analysis script refuses holdout data unless the hash matches. %
The protocol fixed the following:
\begin{itemize}[leftmargin=*,itemsep=1pt,topsep=2pt]
\item the 34 tasks, the three W4 configurations with their weights, and the 12 prompt blocks
\item two fixed halves of six blocks each
\item an outcome-blind repair rule, with reserve blocks used only to repair infrastructure failures
\item the main estimate, the full-pool net gain, with a crossed task$\times$block 90\% interval from two-way analysis-of-variance (ANOVA) moments under a normal approximation
\item the readings of the main estimate, with $L$ the lower interval bound: positive if $L>0$, above 3\pp{} if $L>3$, small if the interval lies inside $\pm$3\pp{}, and inconclusive otherwise
\item two secondary checks: a rescue check, requiring both $r>0$ and $d>0$, and a repeatability check, testing whether task fragility repeats across the two halves with a persona-stratified permutation test (10{,}000 draws, no outcome screen)
\item six stratum families, with Holm correction within each.
\end{itemize}
The readings contain categories for positive and small effects and none for a negative effect, so a negative interval falls under \emph{inconclusive}.

\paragraph{Registered results.} Table~\ref{tab:holdout} gives the main estimate and the per-configuration values. Damages exceeded rescues, and the full-pool net gain was $-4.57$\pp{} with a crossed 90\% interval of [$-7.17$, $-1.96$], entirely below zero, and a $t$ interval over block means gives [$-6.08$, $-3.05$]. The registered reading of this estimate is inconclusive. Both secondary checks passed: rescues and damages were both positive, and task fragility repeated across the two halves (weighted cross-half correlation 0.588). Table~\ref{tab:strata} gives all six stratum families, which the protocol treats as descriptive. Every level estimate is $\le 0$, and the levels that remain significant after Holm correction are negative: step-budget tertiles T2 and T3 (T1 to T3 run from lowest to highest), mobile-data issues ($-9.09$\pp{}) and the fault tertile T1 ($-6.37$\pp{}). %

\begin{table*}[t]
\centering%
\renewcommand{\arraystretch}{1.12}\fontsize{8.5}{10}\selectfont\setlength{\tabcolsep}{4pt}
\begin{tabular}{@{}lrrllll@{}}
\toprule
Configuration (weight) & units & excl. & $\Gall$ & $r$ & $d$ & $\rho_c$ \\
\midrule
4B GPTQ-W4 (1/2) & 396 & 1 & $-3.03$ [$-7.86$, $+1.80$] & $+5.05$ & $+8.08$ & 0.668 \\
8B GPTQ-W4 (1/4) & 408 & 0 & $-5.64$ [$-9.33$, $-1.94$] & $+4.41$ & $+10.05$ & 0.374 \\
8B AWQ-style W4 (1/4) & 396 & 1 & $-6.57$ [$-11.98$, $-1.15$] & $+6.31$ & $+12.88$ & 0.644 \\
\midrule
Weighted ($\Gall$) & & & $-4.57$ [$-7.17$, $-1.96$] & $+5.21$ & $+9.77$ & 0.588 \\
\bottomrule
\end{tabular}
\caption{Telecom holdout, 34 tasks $\times$ 12 blocks (pp of units, crossed 90\% intervals). Units: task--block units analysed. Excl.: tasks structurally excluded because the decision count and the trajectory count disagreed at state search. $\Gall$: full-pool net gain of action-only replacement. $r$, $d$: rescue and damage rates. $\rho_c$: correlation of per-task turnover between the two six-block halves, the statistic of the repeatability check. The weighted row is the registered main estimate.} %
\label{tab:holdout}
\end{table*}

\paragraph{Structural exclusions.} Two task--configuration pairs were excluded as structural (Table~\ref{tab:holdout}), and the complete-case analysis gives the same result. %
A post hoc sensitivity analysis with the registered estimator re-included both pairs. With every one of their units set to $-1$ or $+1$ in all twelve blocks, the point estimate of the full-pool net gain ranged from $-6.68$ [$-10.35$, $-3.01$] to $-2.27$ [$-5.96$, $+1.43$]\pp{} (crossed 90\%), so the interval of the most favourable scenario covers zero. With their observed units kept in the eleven blocks where they exist and only the two missing units set to $-1$ or $+1$, the estimates ranged from $-4.72$ to $-4.35$\pp{}, each with a crossed 90\% interval below zero. The registered reading was inconclusive in all 36 combinations. %

\paragraph{Full replay of the shared baseline.} This check was added after the registered analysis. The QQ outcome comes from the recorded episode, whereas the other branches are re-executed. An exploratory excess of states at which only QQ succeeded, in two 8B configurations, prompted a direct audit of the baseline. An initial hash-selected sample of 135 of the 1202 divergent states matched. An audit protocol fixed before the replay then re-executed all 1202 states, checking the original checkpoint, serving configuration and a numerical canary before each run. Every replay matched the original trajectory, prompts, termination, outcome, budget counters, database and environment state. The recorded and re-executed branches also use the same official success rule. The audit verifies the QQ baseline and leaves open why only QQ succeeds on particular tasks. %

\begin{table*}[t]
\centering%
\renewcommand{\arraystretch}{1.12}\fontsize{8.5}{10}\selectfont\setlength{\tabcolsep}{4pt}
\begin{tabular}{@{}llrlll@{}}
\toprule
Family & Level & $n$ & $\Gall$ [crossed 90\%] & $p$ & Holm $p$ \\
\midrule
step budget$^a$ & T1 & 138/129/110 & $-0.32$ [$-1.57$, $+0.94$] & .6771 & .6771 \\
 & T2 & 124/126/162 & $-2.30$ [$-3.88$, $-0.72$] & .0165 & .0495 \\
 & T3 & 123/144/124 & $-1.94$ [$-3.35$, $-0.54$] & .0229 & .0495 \\
\midrule
persona & Hard & 11 & $-5.99$ [$-12.23$, $+0.26$] & .1147 & .3355 \\
 & None & 12 & $-3.65$ [$-7.42$, $+0.13$] & .1118 & .3355 \\
 & Easy & 11 & $-4.36$ [$-9.19$, $+0.48$] & .1385 & .3355 \\
\midrule
BF16 difficulty & 0 & 20 & $-2.35$ [$-4.09$, $-0.62$] & .0257 & .0772 \\
 & low & 11 & $-4.54$ [$-8.90$, $-0.19$] & .0860 & .1720 \\
 & high & 3 & $-18.75$ [$-38.00$, $+0.50$] & .1091 & .1720 \\
\midrule
task length & T1 & 9 & $-5.09$ [$-9.81$, $-0.38$] & .0755 & .1511 \\
 & T2 & 8 & $-0.78$ [$-4.46$, $+2.90$] & .7270 & .7270 \\
 & T3 & 17 & $-6.17$ [$-10.49$, $-1.84$] & .0189 & .0567 \\
\midrule
issue type & mms & 14 & $-1.15$ [$-3.33$, $+1.04$] & .3880 & .4434 \\
 & mobile data & 11 & $-9.09$ [$-14.98$, $-3.20$] & .0111 & .0333 \\
 & service & 9 & $-4.40$ [$-10.32$, $+1.52$] & .2217 & .4434 \\
\midrule
faults & T1 & 19 & $-6.37$ [$-9.91$, $-2.83$] & .0030 & .0091 \\
 & T2 & 10 & $-1.94$ [$-7.58$, $+3.69$] & .5704 & .5704 \\
 & T3 & 5 & $-2.92$ [$-5.18$, $-0.66$] & .0337 & .0674 \\
\bottomrule
\end{tabular}
\caption{Telecom holdout strata: the six registered families, with Holm correction within each family over its three levels (pp). T1--T3 are tertiles from lowest to highest, with cut points fixed on telecom development tasks. For task-level families, $n$ is the number of tasks and $\Gall$ is computed over the tasks in the level. $^a$The step budget (remaining steps at the divergence) is a unit-level family: $n$ gives the number of divergent units per configuration (4B GPTQ-W4/8B GPTQ-W4/8B AWQ-style W4), and $\Gall$ is the level's contribution to the overall $\Gall$, so the three levels sum to $\Gall$.}
\label{tab:strata}
\end{table*}

\subsection{Telecom development tasks}
\label{app:holdoutdev}

The holdout changed domain as well as tasks. The main development panel is retail and airline, and the holdout tasks are all telecom. On 80 telecom development tasks (six blocks, fixed-pool 95\% intervals), the two 8B holdout configurations gave full-pool net gains of $+2.7$ [$-4.7$, $+10.1$] (GPTQ-W4) and $-0.6$ [$-4.9$, $+3.7$]\pp{} (AWQ-style W4), against $-5.6$ and $-6.6$\pp{} on the holdout. Qwen3-4B was not forked on telecom development tasks. With six-block intervals this wide, a telecom domain effect and heterogeneity among the holdout tasks remain confounded. %
\section{Scope Across States, Environments and Models}
\label{app:scope}

This appendix supports the scope results in \S\ref{sec:value}: ordinary states, sensitivity to the user simulator, a second model family and environments without an LLM user. W4 and W3 denote 4- and 3-bit weights, KV8 and KV4 8- and 4-bit KV caches, and RTN round-to-nearest quantization. The five stress settings of Table~\ref{tab:scope} are expanded with all intervals in Table~\ref{tab:ghf}. %

Table~\ref{tab:settings} summarises the action-only net gain across all settings.

\begin{table}[t]
\centering\small
\setlength{\tabcolsep}{3pt}\renewcommand{\arraystretch}{1.12} %
\begin{tabular*}{\columnwidth}{@{\extracolsep{\fill}}llr@{}}
\toprule
\textbf{Setting} & \textbf{States} & \textbf{Net gain [CI]} \\
\midrule
\tautwo{} retail, airline & all units & $+0.7$ [$-2.1$, $+3.4$] \\
\tautwo{} telecom holdout & all units & $-4.6$ [$-7.2$, $-2.0$] \\
BFCL multi-turn & all items & $+0.1$ [$-1.0$, $+1.2$] \\
ALFWorld (8B) & all items & $-3.7$ [$-9.7$, $+1.9$] \\
\addlinespace[2pt] %
Mistral GPTQ-W4 & ordinary & $-0.8$ [$-5.2$, $+3.6$] \\
Mistral RTN-W4 & ordinary & $+0.7$ [$-2.7$, $+4.2$] \\
\bottomrule
\end{tabular*}
\caption{Action-only net gain (pp) across settings. Qwen3 rows are W4 pools over all units or items, with units without a divergence counted as zero. Mistral rows are $G$ at its own ordinary states on 83 retail and airline tasks. Intervals are new-task 95\% (retail, airline), crossed 90\% (holdout), item-bootstrap 95\% (BFCL, ALFWorld) and fixed-pool 95\% (Mistral). Per-configuration values: Appendix~\ref{app:extr2} and~\ref{app:mistral}.}
\label{tab:settings}
\end{table}

\subsection{Ordinary states and the user simulator}
\label{app:ordinary}

\paragraph{Turnover at ordinary states.} At ordinary states, replacing the quantized action changed the outcome less often than at first divergent states (a descriptive comparison of different state populations, reported without an interval on the difference). States were drawn from three history sources: the quantized model's own histories, BF16 histories shared by all configurations of a model, and the histories of one quantized configuration shared by all configurations of a model. Over the first three blocks (task-equal, $t$ intervals over block means), turnover $M$ was 7.8--9.0\% for the Qwen3-4B W4 configurations on states from the quantized model's own and from BF16 histories, 4.7--9.0\% for the Qwen3-8B W4 configurations and at most 4.7\% for KV8. The net gain lay within $\pm$2.8\pp{} on states from the quantized model's own and from BF16 histories and within $\pm$5.0\pp{} on states from the shared quantized histories, and every interval covers zero. Ordinary-state analyses use two to six blocks, so three-block intervals rest on two degrees of freedom. %
Token-identical actions preserve the outcome under deterministic serving. Reused arms were first selected by canonical equality and then checked with each model's chat template, and 4{,}113 of 4{,}114 reuses were token-identical. The remaining pair differs only in JSON key order. Its two cross arms were measured separately and reproduced the reused outcomes. %

\paragraph{Sensitivity of text-difference flips to the user simulator.} Text actions act through the user simulator, and changing the user LLM moved agent success on $\tau$-bench retail by up to 9 percentage points \citep{seshadri2026lost} and mean task reward on $\tau$-bench-family enterprise tasks by 15.2 points \citep{jain2026userproxybench}. %
The control was fixed before it was run. It took the 39 deduplicated text-difference flips with the same tool structure from the first block and 39 matched non-flip states, and re-ran their BQ and QQ branches three times each with a different user simulator, Sonnet 5.5, sampled with its default settings and without temperature or seed control, so neither simulator control can be repeated draw for draw. The simulator answered through a file relay served by Claude Code subagents. Subagents answered many relay requests in one context, so a responder could see other arms or replicates of the same conversation (further reproducibility limits in the Limitations). %
Judged by the sign of the replicate mean, 7 of the 39 flips kept their direction (95\% CI 0.075--0.335), 6 reversed and 26 vanished (15 both succeed, 2 both fail, 9 ties). Among the matched non-flip states, 11 of 39 became flips (0.15--0.45). Replicates of the same arm disagreed in 41\% of flip arms and in 21--26\% of control arms. %
Most text-difference labels changed under this alternative setting. Simulator model, sampling and relay interface changed together, so their separate effects remain unidentified.

\paragraph{Sensitivity of the four-branch results.}
\label{app:simctrlpmain}
Before running, we sampled 80 first divergent states without outcome information (20 per W4 configuration, stratified by domain and block) and re-ran all four branches five times each with Sonnet 5.5 as the user. Each user turn was one isolated single-request call (7896 calls), which removed the cross-arm exposure of the relay above, and all integrity checks passed. %
Per-state branch outcomes tracked the original single runs with correlation 0.63--0.68. Pooled over configurations, Sonnet gave action replacement $G=-3.0$ [$-10.2$, $+4.8$], takeover $H=+5.5$ [$+0.0$, $+11.2$] and a difference $G-H=-8.5$ [$-14.8$, $-2.8$]\pp{}, against $G-H=-2.5$ [$-11.2$, $+6.2$] on the same states under the original simulator (change $-6.0$ [$-16.2$, $+4.2$]). %
With 20 states per configuration, this control has too few states to resolve the few-point contrasts of Table~\ref{tab:ghf}. On this sample the original simulator itself gives no difference between action replacement and takeover ($G-H=0.0$) in both Qwen3-8B W4 configurations, so the within-configuration orderings stay unresolved under this control, neither confirmed nor reversed. %

\subsection{Models compared on their own ordinary states}
\label{app:mistral}
\paragraph{Mistral population, scoring and coverage.} The Mistral panel applies the same row rule to Mistral's own BF16 trajectories on the 85 retail and airline tasks. Seven BF16 source rows are missing because the source episode ended in an invalid tool call produced by the model itself (airline task 44 in all six blocks, airline task 13 in the first block). A dated amendment, fixed from the BF16 source runs before any Mistral fork output existed, excludes both tasks in every block and configuration, leaving 83 tasks. The selection was outcome-blind, but BF16 Mistral fails these tasks structurally, so the estimates describe this conditional population rather than the 85-task panel and are kept separate from the Qwen3-8B estimates. %
A branch that ended because the harness could not accept the agent's own message (an empty or invalid tool name, or an empty message) is scored as a failure, as in end-to-end evaluation. This scoring rule was adopted after the results of the first phase had been read. Without it, the first phase (one complete block) left the registered GPTQ $-$ RTN comparison inestimable. Two sensitivity analyses are reported alongside. One omits the rule, which leaves such an arm missing and its block incomplete. The other applies the rule only to data still unread when the rule was fixed. Both admit the same three blocks for GPTQ $-$ RTN ($+1.2$ [$-13.8$, $+16.2$]). Coverage: GPTQ-W4 has 498 states on six blocks and RTN-W4 415 states on five. %
\paragraph{Estimates.} On the 83 tasks (GPTQ-W4 on six blocks, RTN-W4 on five, Table~\ref{tab:mistral-modes}), the BF16 action in GPTQ-W4 rescued $r=4.6$ [2.4, 6.8] and damaged $d=5.4$ [2.4, 8.5]\pp{} of states ($M=10.0$ [7.0, 13.1], fixed-pool 95\%), and every mode's net value was unresolved ($G=-0.8$ [$-5.2$, $+3.6$], $H=+1.4$ [$-3.0$, $+5.8$], $F=+0.0$ [$-4.2$, $+4.2$]). RTN-W4 was similar ($M=10.4$ [6.3, 14.4]), and all three of its mode intervals also cover zero. The pre-registered contrast GPTQ $-$ RTN in $D_{cc}=\YY{QB}-\YY{BB}$ was unresolved: $\Delta D_{cc}=+2.2$ [$-3.5$, $+7.8$]\pp{}, $p=.35$, on five blocks, two of which are admitted only by the scoring rule adopted after the interim look. %

\begin{table*}[t]
\centering%
\renewcommand{\arraystretch}{1.12}\fontsize{8.5}{10}\selectfont\setlength{\tabcolsep}{4pt}
\begin{tabular}{@{}lrrlll@{}}
\toprule
Configuration & blocks & states & $G$ & $H$ & $F$ \\
\midrule
GPTQ-W4 & 6 & 498 & $-0.8$ [$-5.2$, $+3.6$] & $+1.4$ [$-3.0$, $+5.8$] & $+0.0$ [$-4.2$, $+4.2$] \\
 & & & [$-5.8$, $+4.2$] & [$-4.0$, $+6.8$] & [$-6.0$, $+6.0$] \\
RTN-W4 & 5 & 415 & $+0.7$ [$-2.7$, $+4.2$] & $+1.9$ [$-3.8$, $+7.7$] & $+2.4$ [$-2.2$, $+7.0$] \\
 & & & [$-3.6$, $+5.5$] & [$-4.6$, $+8.0$] & [$-3.9$, $+8.4$] \\

\bottomrule
\end{tabular}
\caption{Mistral-Small-3.2-24B on its fixed 83-task conditional population of ordinary states sampled from BF16 histories. Effects in pp. Each entry gives the estimate with its fixed-pool 95\% $t$ interval, then the crossed task$\times$block 95\% interval. RTN-W4's fourth block is excluded because of missing argument-JSON errors, while model-produced invalid tool names are scored as failures under the dated amendment (Appendix~\ref{app:mistral}).}
\label{tab:mistral-modes}
\end{table*}

\subsection{Upgrade effects on BFCL and ALFWorld}
\label{app:extr2}

Table~\ref{tab:extr2} gives the replacement quantities of \S\ref{sec:value} ($r$, $d$, $M$ and $\Gall$ over all items, non-divergent items counting as zero) and applies the four-branch analysis to the external benchmarks, from the raw four-arm outcomes on divergent items (one run per item, item bootstrap within benchmark category). The four-branch columns are post hoc analyses outside any pre-registered plan. In BFCL, Qwen3-4B AWQ-style W4 had a negative net gain, $\Gall=-3.8$ [$-6.5$, $-1.3$]\pp{}, with action replacement at $G=-4.7$ [$-7.8$, $-1.5$]\pp{} on its divergent states. In the unrotated KV4 configurations, $\YY{QQ}=0$ on every state, so all positive upgrades are rescues. Action-only replacement stayed near zero in these configurations, while continuation and joint upgrades helped. In BFCL Qwen3-4B GPTQ-W3, the QQ no-op rule dropped the base category whole, leaving 591 items and 540 states. %
Off-diagonal shares ($\DQ\neq\DB$) were 4.7--17.1\% in the BFCL W4 configurations (pool 10.3\%) and 32.8--40.0\% in Qwen3-8B ALFWorld W4 (pool 36.4\%). The ALFWorld GPTQ-W3 stress configuration had takeover $H=+38.5$ and a difference between action replacement and takeover of $G-H=-34.6$ [$-43.5$, $-25.4$]\pp{}, and the unrotated KV4 rows also separate continuation takeover from joint replacement. %

\begin{table*}[t]
\centering%
\renewcommand{\arraystretch}{1.12}\fontsize{8.5}{10}\selectfont\setlength{\tabcolsep}{4pt}
\begin{tabular*}{\textwidth}{@{\extracolsep{\fill}}llllll@{}}
\toprule
\multicolumn{6}{@{}l}{\textbf{A. Full item pool}} \\
\addlinespace[2pt]
Configuration & items & $r$ & $d$ & $M$ & $\Gall$ \\
\midrule
BFCL 4B GPTQ-W4 & 788 & 6.3 & 4.3 & 10.7 & $+2.0$ [$-0.3$, $+4.3$] \\
BFCL 4B AWQ-style W4 & 788 & 5.7 & 9.5 & 15.2 & $-3.8$ [$-6.5$, $-1.3$] \\
BFCL 8B GPTQ-W4 & 788 & 2.9 & 2.3 & 5.2 & $+0.6$ [$-0.9$, $+2.3$] \\
BFCL 8B AWQ-style W4 & 788 & 4.1 & 2.7 & 6.7 & $+1.4$ [$-0.3$, $+3.3$] \\
BFCL W4 pool &  &  &  &  & $+0.1$ [$-1.0$, $+1.2$] \\
ALFWorld 8B GPTQ-W4 & 134 & 9.0 & 12.7 & 21.6 & $-3.7$ [$-11.2$, $+4.5$] \\
ALFWorld 8B AWQ-style W4 & 134 & 10.4 & 14.2 & 24.6 & $-3.7$ [$-11.9$, $+4.5$] \\
ALFWorld W4 pool &  &  &  &  & $-3.7$ [$-9.7$, $+1.9$] \\
\midrule
BFCL 4B GPTQ-W3$^s$ & 591 & 6.1 & 3.6 & 9.6 & $+2.5$ [$+0.2$, $+5.1$] \\
BFCL 8B GPTQ-W3$^s$ & 788 & 5.2 & 4.1 & 9.3 & $+1.1$ [$-1.0$, $+3.3$] \\
BFCL 4B KV4$^s$ & 788 & 0.3 & 0.0 & 0.3 & $+0.3$ [$+0.0$, $+0.6$] \\
BFCL 8B KV4$^s$ & 788 & 0.3 & 0.0 & 0.3 & $+0.3$ [$+0.0$, $+0.6$] \\
ALFWorld 8B GPTQ-W3$^s$ & 134 & 6.7 & 3.0 & 9.7 & $+3.7$ [$-1.5$, $+9.7$] \\
ALFWorld 8B KV4$^s$ & 134 & 0.0 & 0.0 & 0.0 & $+0.0$ [$+0.0$, $+0.0$] \\
\bottomrule
\end{tabular*}
\par\vspace{6pt}
\begin{tabular*}{\textwidth}{@{\extracolsep{\fill}}lllllll@{}}
\toprule
\multicolumn{7}{@{}l}{\textbf{B. First divergent states}} \\
\addlinespace[2pt]
Configuration & states & $G$ & $H$ & $F$ & $G-H$ & off-diag.\ \% \\
\midrule
BFCL 4B GPTQ-W4 & 573 & \shortstack[l]{$+2.8$\\{}[$-0.4$, $+5.9$]} & \shortstack[l]{$+1.4$\\{}[$-0.7$, $+3.5$]} & \shortstack[l]{$+4.2$\\{}[$+0.9$, $+7.2$]} & \shortstack[l]{$+1.4$\\{}[$-1.8$, $+4.5$]} & 11.7 \\
BFCL 4B AWQ-style W4 & 642 & \shortstack[l]{$-4.7$\\{}[$-7.8$, $-1.5$]} & \shortstack[l]{$-1.1$\\{}[$-3.6$, $+1.4$]} & \shortstack[l]{$-1.4$\\{}[$-4.8$, $+1.7$]} & \shortstack[l]{$-3.6$\\{}[$-6.7$, $-0.2$]} & 17.1 \\
BFCL 8B GPTQ-W4 & 533 & \shortstack[l]{$+0.9$\\{}[$-1.3$, $+3.3$]} & \shortstack[l]{$+1.5$\\{}[$+0.0$, $+3.2$]} & \shortstack[l]{$+2.1$\\{}[$-0.4$, $+4.6$]} & \shortstack[l]{$-0.6$\\{}[$-3.1$, $+2.0$]} & 4.7 \\
BFCL 8B AWQ-style W4 & 548 & \shortstack[l]{$+2.0$\\{}[$-0.4$, $+4.7$]} & \shortstack[l]{$+1.8$\\{}[$+0.0$, $+3.7$]} & \shortstack[l]{$+3.6$\\{}[$+0.9$, $+6.5$]} & \shortstack[l]{$+0.2$\\{}[$-2.6$, $+3.1$]} & 7.8 \\
BFCL W4 pool &  & \shortstack[l]{$+0.3$\\{}[$-1.2$, $+1.8$]} & \shortstack[l]{$+0.9$\\{}[$-0.2$, $+2.0$]} & \shortstack[l]{$+2.1$\\{}[$+0.3$, $+3.9$]} & \shortstack[l]{$-0.6$\\{}[$-2.3$, $+1.0$]} & 10.3 \\
ALFWorld 8B GPTQ-W4 & 115 & \shortstack[l]{$-4.3$\\{}[$-13.3$, $+5.0$]} & \shortstack[l]{$+7.0$\\{}[$-2.6$, $+16.7$]} & \shortstack[l]{$+9.6$\\{}[$-0.9$, $+19.7$]} & \shortstack[l]{$-11.3$\\{}[$-20.9$, $-1.8$]} & 40.0 \\
ALFWorld 8B AWQ-style W4 & 119 & \shortstack[l]{$-4.2$\\{}[$-13.6$, $+4.9$]} & \shortstack[l]{$+5.0$\\{}[$-2.6$, $+12.2$]} & \shortstack[l]{$+4.2$\\{}[$-5.9$, $+13.7$]} & \shortstack[l]{$-9.2$\\{}[$-18.8$, $+0.0$]} & 32.8 \\
ALFWorld W4 pool &  & \shortstack[l]{$-4.3$\\{}[$-11.1$, $+2.2$]} & \shortstack[l]{$+6.0$\\{}[$-0.5$, $+12.6$]} & \shortstack[l]{$+6.9$\\{}[$-2.1$, $+15.5$]} & \shortstack[l]{$-10.3$\\{}[$-17.7$, $-2.9$]} & 36.4 \\
\midrule
BFCL 4B GPTQ-W3$^s$ & 540 & \shortstack[l]{$+2.8$\\{}[$+0.2$, $+5.5$]} & \shortstack[l]{$+5.2$\\{}[$+2.9$, $+7.7$]} & \shortstack[l]{$+8.9$\\{}[$+5.7$, $+12.2$]} & \shortstack[l]{$-2.4$\\{}[$-5.7$, $+0.7$]} & 13.0 \\
BFCL 8B GPTQ-W3$^s$ & 696 & \shortstack[l]{$+1.3$\\{}[$-1.1$, $+3.7$]} & \shortstack[l]{$+1.4$\\{}[$-0.1$, $+2.9$]} & \shortstack[l]{$+3.4$\\{}[$+0.9$, $+6.0$]} & \shortstack[l]{$-0.1$\\{}[$-2.6$, $+2.4$]} & 8.3 \\
BFCL 4B KV4$^s$ & 787 & \shortstack[l]{$+0.3$\\{}[$+0.0$, $+0.6$]} & \shortstack[l]{$+2.9$\\{}[$+1.9$, $+4.1$]} & \shortstack[l]{$+20.3$\\{}[$+17.6$, $+23.0$]} & \shortstack[l]{$-2.7$\\{}[$-3.9$, $-1.5$]} & 20.5 \\
BFCL 8B KV4$^s$ & 788 & \shortstack[l]{$+0.3$\\{}[$+0.0$, $+0.6$]} & \shortstack[l]{$+1.8$\\{}[$+1.0$, $+2.7$]} & \shortstack[l]{$+14.5$\\{}[$+12.1$, $+17.0$]} & \shortstack[l]{$-1.5$\\{}[$-2.5$, $-0.6$]} & 14.7 \\
ALFWorld 8B GPTQ-W3$^s$ & 130 & \shortstack[l]{$+3.8$\\{}[$-1.5$, $+9.8$]} & \shortstack[l]{$+38.5$\\{}[$+29.8$, $+46.9$]} & \shortstack[l]{$+42.3$\\{}[$+33.6$, $+50.8$]} & \shortstack[l]{$-34.6$\\{}[$-43.5$, $-25.4$]} & 30.8 \\
ALFWorld 8B KV4$^s$ & 134 & \shortstack[l]{$+0.0$\\{}[$+0.0$, $+0.0$]} & \shortstack[l]{$+38.1$\\{}[$+30.6$, $+45.5$]} & \shortstack[l]{$+53.0$\\{}[$+44.8$, $+61.2$]} & \shortstack[l]{$-38.1$\\{}[$-45.5$, $-30.6$]} & 26.9 \\
\bottomrule
\end{tabular*}
\par\vspace{6pt}
\caption{Replacement and four-branch quantities on BFCL multi-turn and ALFWorld (no LLM user simulator, one run per item, 95\% item-bootstrap intervals). 4B and 8B denote Qwen3-4B and Qwen3-8B. Panel A, all items: rescue $r$, damage $d$, turnover $M$ (\% of all items) and net gain $\Gall$ (pp), non-divergent items counting as zero. Panel B, divergent items (post hoc, pp): action $G$, continuation $H$, joint $F$, and the off-diagonal share, the percentage of states with $\DQ\neq\DB$. Pools weight configurations equally. $^s$Stress and unrotated KV4 configurations. In unrotated KV4 rows, $\YY{QQ}=0$ on every state, so every positive gain is a rescue.}
\label{tab:extr2}
\end{table*}

\section{Precision References and Common-State Comparisons}
\label{app:sameprec}

This appendix supports \S\ref{sec:sameprec}. The second-checkpoint reference covers GPTQ checkpoints whose calibration samples were redrawn from the same calibration trajectories, and the AWQ-style, KV8, RTN and rotated KV4 settings have no such reference. This appendix gives the full results for the second checkpoint, the common-state comparison of the BF16 action with the second checkpoint's, and the key-order control. Table~\ref{tab:refsfull} collects the reference comparisons.

\begin{table*}[t]
\centering%
\renewcommand{\arraystretch}{1.12}\fontsize{8.5}{10}\selectfont\setlength{\tabcolsep}{4pt}
\begin{tabular}{@{}P{2.1cm}P{2.9cm}P{1.8cm}rrrP{4.7cm}@{}}
\toprule
\raggedright Reference & \raggedright States (denominator) & \raggedright Action replacing Q's (continuation) & \begin{tabular}[b]{@{}r@{}}Turn-\\over\end{tabular} & \begin{tabular}[b]{@{}r@{}}Rescues /\\damages\end{tabular} & Net & \raggedright Paired comparison [interval; 95\% unless marked] \tabularnewline
\midrule
\raggedright Second W4 checkpoint Q$'$ & \raggedright Own first divergences; Qwen3-4B and Qwen3-8B GPTQ-W4; 1020 task--block units (85 tasks $\times$ 6 blocks per model) & \raggedright B (Q) & 16.8 & 8.0 / 8.7 & $-0.7$ & \raggedright $\Delta M=-1.0$; fixed [$-3.9$, $+1.9$]; new [$-4.5$, $+2.7$] \tabularnewline %
 & & \raggedright Q$'$ (Q) & 17.7 & 9.3 / 8.4 & $+0.9$ & \tabularnewline
\midrule
\raggedright Common states & \raggedright 510 BF16-sourced Qwen3-8B states; 85 tasks $\times$ 6 blocks; GPTQ-W4 substitutes & \raggedright B (Q) & 7.1 & 4.1 / 2.9 & $+1.2$ & \raggedright $K(\mathrm{Q})=+1.0$; two-look 97\% $t$ [$-4.2$, $+6.2$]; bootstrap [$-1.4$, $+3.5$] \tabularnewline %
 & & \raggedright Q$'$ (Q) & 8.0 & 4.1 / 3.9 & $+0.2$ & \tabularnewline
\addlinespace
 & & \raggedright B (B; vs.\ QB) & 9.0 & 5.3 / 3.7 & $+1.6$ & \raggedright $K(\mathrm{B})=+0.8$; two-look 97\% $t$ [$-2.2$, $+3.8$]; bootstrap [$-2.0$, $+3.7$] \tabularnewline
 & & \raggedright Q$'$ (B; vs.\ QB) & 8.6 & 4.7 / 3.9 & $+0.8$ & \tabularnewline
\midrule
\raggedright Key order & \raggedright 300 Qwen3-8B GPTQ-W4 first divergent states whose action is one tool call with at least two argument keys, 41 tasks (\% of states) & \raggedright B (Q) & 9.0 & 3.3 / 5.7 & $-2.3$ & \raggedright turnover, B minus reversed keys: $+2.3$ [$-2.9$, $+8.4$] \tabularnewline
 & & \raggedright Q, keys reversed (Q) & 6.7 & 3.3 / 3.3 & $0.0$ & \tabularnewline
\bottomrule
\end{tabular}
\caption{Reference comparisons behind Figure~\ref{fig:refs} and the common-state paragraph of \S\ref{sec:sameprec}, in percentage points (pp). Each row group has its own state source and denominator, and comparisons are paired only within a group. In the table, B denotes the BF16 model, Q the quantized model and Q$'$ the second checkpoint. The references are reported beside the BF16 turnover and left unsubtracted. Turnover, rescues, damages and net gain compare the replacing action with the quantized action under the continuation shown in parentheses: against QQ when the quantized model continues, and against the quantized action followed by the BF16 model (QB) when the BF16 model continues. $\Delta M$ is the BF16 turnover minus the second checkpoint's, and $K(c)$ is success with the BF16 action minus success with the second checkpoint's action under continuation $c$. Intervals follow each protocol. For the second checkpoint, ``fixed'' is a $t$ interval over 6 block means and ``new'' a crossed new-task bootstrap. For common states, a reading needs both the $t$ interval over block means and the task-cluster bootstrap, shown here under the two-look (Pocock) reading that the protocol designates confirmatory. For key order, the interval is the wider of block and crossed bootstrap. The key-order control is exploratory, its analysis code was frozen before outcomes were read, and it set no equivalence margin. Per-arm intervals appear in Appendices~\ref{app:qprime} and~\ref{app:keyorder}, and unadjusted common-state intervals and the extension timeline in Appendix~\ref{app:commonstate}.}
\label{tab:refsfull}
\end{table*}

\subsection{Second same-bit-width checkpoint}
\label{app:qprime}

\paragraph{Replay checks.} Every state in both designs passed replay of the quantized model, and every branch passed the prefix and environment checks. QQ no-op samples passed 265/265 in the BF16 design and 242/242 in the second-checkpoint design, and no run directory was quarantined. %

\paragraph{Equivalence tests (W4).} Pooled over Qwen3-4B and Qwen3-8B, the turnover difference was $\Delta M=M-M'=-0.98$\pp{} (90\% CI [$-3.26$, $+1.30]$). The comparison detected no precision-specific excess, and equivalence within $\pm$3\pp{} also stayed unresolved (two one-sided tests, $p=0.067$). The net gain difference behaved the same way ($\Delta\Gall=-1.57$ [$-5.17$, $+2.04$], $p=0.230$), as did, descriptively, the difference in off-diagonal share ($-1.210$ [$-4.09$, $+1.67$], $p=0.133$). The interval and test for the off-diagonal share use the mean of per-block ratios, and the ratio of block sums gives $-1.216$. %
The protocol first read ``interval includes 0 and $|\Delta M|\le3$\pp{}'' as similar turnover. After the interim result had been seen, a dated amendment replaced that reading with the stricter two one-sided tests and left every estimate unchanged.

\paragraph{Per model.} On the final analysis set ($n_{\mathrm{blk}}=6$), the turnover difference was $\Delta M=+0.6$ [$-3.1$, $+4.3$]\pp{} for Qwen3-4B and $-2.5$ [$-5.6$, $+0.5$] for Qwen3-8B (fixed pool, 95\%), with new-task intervals [$-5.5$, $+6.9$] and [$-7.3$, $+2.7$]. Equivalence within $\pm$3\pp{} stayed unresolved for every per-model difference in turnover and in full-pool net gain. The Qwen3-8B off-diagonal difference ($+0.40$, 90\% [$-2.12$, $+2.92$]) is the only per-model contrast inside the margin. %

\paragraph{Full substitution (W4).} When the substituted model also continued, the BF16 model flipped more units than the second checkpoint (193 against 149, $+4.3$ [$+0.4$, $+8.3$]\pp{}, unadjusted for the several contrasts reported here). This is a turnover comparison and is separate from the unresolved action-value comparison above. %

\paragraph{W3 stress pair.} The W3 panel uses GPTQ-W3 as the quantized model and a second W3 calibration draw as the second checkpoint, on Qwen3-4B and Qwen3-8B with three blocks (510 units). It is reported beside W4 and kept unpooled. The same-precision substitution produced 17.3\pp{} of turnover and the BF16 substitution 18.8\pp{}, with off-diagonal shares of 27.6\% and 27.3\%. %
The turnover difference $\Delta M=+1.6$\pp{} (95\% [$-0.66$, $+3.80$]) is unresolved. The registered $\pm$3\pp{} equivalence test gave $p=0.055$, and the 90\% interval lies just above zero while the registered reading uses the 95\% interval. The difference in full-pool net gain ($\Delta\Gall$) is equivalent within $\pm$3\pp{} only in the registered pooled population ($+1.2$, $p=0.017$). Per model it is $+3.1$ (Qwen3-4B) and $-0.8$\pp{} (Qwen3-8B). The off-diagonal difference is $-0.288$ as a ratio of block sums and $-0.307$ as a mean of per-block ratios, which the interval and test use (90\% [$-11.0$, $+10.4]$, $p=0.27$). %
The W3 run passed the same checks. Every state passed replay of the quantized model, QQ no-op samples passed 31/31 and 39/39, and no run directory was quarantined. All branches passed the prefix and environment checks except one of 774 in the Qwen3-8B same-precision design, a Q$'$Q$'$ branch that ended in an agent-side context overflow. The harness recorded no prefix comparison for that branch. Its environment hash and first replayed prompt match the frozen state, and the protocol counts it as a task failure and keeps it in the analysis. %

\subsection{Common states: additional value of the BF16 action}
\label{app:commonstate}

\paragraph{Design.} In the comparisons above each substitution acts at its own first divergent states. The common-state design removes that difference. On 510 ordinary BF16-sourced Qwen3-8B states (85 tasks $\times$ 6 blocks, sampled by a fixed rule), the current action comes from the BF16 model, from the quantized model (GPTQ-W4, first checkpoint) or from the second checkpoint, and the BF16 or the quantized model continues. The estimand $K(c)=V(\mathrm{B},c)-V(\mathrm{Q}',c)$ measures what the BF16 action adds beyond another W4 action under continuation $c$. The frozen reading rule claims value only if both the block-$t$ and the task-bootstrap interval exclude zero (Holm), and equivalence only if both lie within $\pm$3\pp{} (two one-sided tests). %

\paragraph{Extension after the first look.} The design was registered at three blocks, and that first look was unresolved. With the quantized model continuing, $K=-0.4$\pp{} (95\%: $t$ [$-9.3$, $+8.5$], bootstrap [$-3.5$, $+2.7$]), and with the BF16 model continuing, $K=+2.4$\pp{} ([$-3.5$, $+8.2$], [$-0.8$, $+5.5$]). Three more blocks were then added. A dated amendment fixed the extension and a two-look reading rule after that first look and before any extension run had produced output, and the protocol designates the two-look reading as confirmatory. It uses Pocock levels from a normal approximation, about 97\% intervals for a difference and 94\% for equivalence. The Pocock constant is calibrated for normal statistics, so for small-$k$ $t$ intervals, the task bootstrap and the joint rule its error control is approximate. Because the extension was decided after the first look, the two-look reading also controls the error rate only partly when conditioned on that look. We report it beside the unadjusted reading at the frozen levels and claim an advantage or equivalence only when both readings agree. %

\paragraph{Six-block results.} After the extension, both an advantage of the BF16 action over the second checkpoint's and their equivalence remained unresolved. With the quantized model continuing, $K=+1.0$\pp{} (95\%: $t$ [$-3.5$, $+5.4$], bootstrap [$-1.2$, $+3.1$], and two-look 97\%: [$-4.2$, $+6.2$], [$-1.4$, $+3.5$]). With the BF16 model continuing, $K=+0.8$\pp{} (95\%: [$-1.8$, $+3.3$] for both, and two-look 97\%: [$-2.2$, $+3.8$], [$-2.0$, $+3.7$]). For equivalence with the BF16 model continuing, the unadjusted 90\% intervals lie within $\pm$3\pp{} ($t$ [$-1.2$, $+2.8$], bootstrap [$-1.4$, $+2.9$]), while the 94\% interval of the confirmatory two-look reading extends past the margin ($t$ [$-1.6$, $+3.2$]). With the two readings in disagreement and the extension decided after the first look, equivalence stays unclaimed. %
Outcomes at 26 of the states had already been produced by earlier studies. Without them, $K$ is $+0.5$ and $+0.8$\pp{}, both unresolved, and the unadjusted reading with the BF16 model continuing also leaves equivalence unresolved. BF16's own outcome at the extension states was known in aggregate before the amendment.

\paragraph{Components and action agreement.} With the quantized model continuing, the BF16 action flipped $M=7.1$\% of outcomes (rescues 4.1, damages 2.9) for a net gain of $+1.2$\pp{} (95\%: $t$ [$-1.3$, $+3.6$]). The second checkpoint's action flipped 8.0\% (4.1 and 3.9) for $+0.2$\pp{} ([$-2.3$, $+2.7$]), within $\pm$3\pp{} by two one-sided tests. This auxiliary equivalence compares the second checkpoint with the quantized model, and the comparison of the BF16 model with the second checkpoint is $K$ above. The second checkpoint's action departed from the quantized action at least as often as the BF16 action did: the second checkpoint agreed with the quantized action at 53.9\% of the states and the BF16 model at 55.7\%, and the BF16 model agreed with the second checkpoint at 57.3\%. The interaction of $K$ with the continuation is $-0.2$\pp{} (95\%: $t$ [$-4.5$, $+4.1$], bootstrap [$-2.7$, $+2.4$]), a secondary contrast that is unresolved. %

\paragraph{Qwen3-4B.} The same design on Qwen3-4B was a separate prospective analysis planned at three blocks. The frozen quarantine rule removed the first block (identity checks below), which left two blocks and 170 states. $K=-1.2$\pp{} with the quantized model continuing (bootstrap 95\% [$-5.9$, $+3.5$]) and $-2.4$\pp{} with the BF16 model continuing ([$-6.5$, $+1.2$]), both unresolved. With two blocks the $t$ interval has one degree of freedom. With the quantized model continuing, the two block means happened to be equal, so the $t$ interval collapses to a point. That point reflects two coinciding values rather than a precise measurement, and the reading rests on the bootstrap interval. %

\paragraph{Identity checks.} Across both models, 1{,}238 of 1{,}240 reference and implicit identity checks passed. Both failures came from one Qwen3-4B state in the first block. There the substituted action was semantically identical to the original after parsing, but its token sequence placed the keys of a tool-call argument in a different order. The canonical comparison treats such actions as equal, while the chat template renders the key order, so the next prompt differed and the continuation departed from a replay. Every check with a token-identical action passed (869/869). As the frozen rule requires, the block was removed. %

\subsection{Estimation precision}
\label{app:resolution}

Table~\ref{tab:resolution} lists the intervals of the main unresolved benefit comparisons as reported above. They show how finely each comparison was resolved and are distinct from minimum detectable effects or a retrospective power calculation. Under both continuations, the common-state intervals extend past the $\pm$3\pp{} margin, so a BF16 benefit of a few points remains possible. The 510 rows are 85 tasks repeated across six blocks, and adding blocks and adding tasks reduce different sources of variation, so a prospective power analysis would have to keep that crossed design and specify the target population, effect size and decision rule. %

\begin{table}[ht]
\centering\small
\setlength{\tabcolsep}{4pt}\renewcommand{\arraystretch}{1.12} %
\begin{tabular*}{\columnwidth}{@{\extracolsep{\fill}}lr@{}}
\toprule
Benefit comparison & Estimate [interval] (pp) \\
\midrule
W4 action-only vs.\ QQ & $+0.7$ [$-2.1$, $+3.4$] \\
BF16 vs.\ Q$'$, Q continues & $+1.0$ [$-4.2$, $+6.2$] \\
BF16 vs.\ Q$'$, B continues & $+0.8$ [$-2.2$, $+3.8$] \\
\bottomrule
\end{tabular*}
\caption{Intervals of three benefit comparisons, in percentage points (pp). In row labels, B is the BF16 model, Q the quantized model and Q$'$ the second checkpoint. W4 row: net gain of the BF16 action with the quantized model continuing, against QQ, with the full-pool new-task 95\% interval. Common-state rows: $K(c)$ with the two-look 97\% $t$ interval over block means. Bootstrap intervals appear in Appendix~\ref{app:commonstate}. Rows differ in population and confidence level.}
\label{tab:resolution}
\end{table}

\subsection{Key-order control}
\label{app:keyorder}

\paragraph{Design.} This control is exploratory. Eligible states are first divergent states of Qwen3-8B GPTQ-W4 at which the quantized action is a tool call with at least two top-level argument keys, reversing the keys leaves the parsed call and its environment effects unchanged, and the next prompt tokens differ. This gave 300 of 1473 divergent states from 41 tasks. In the branch KQ, the reversed-key action is forced and the quantized model continues, and QQ and BQ are reused from the first-divergence runs. Eligibility was fixed without outcomes, and the analysis code was frozen, with recorded hashes, before any KQ outcome was read. All four run jobs passed acceptance (KQ checks 300/300, shadow checks 300/300, QQ no-op 35/35, with no errors or context overflows). Two of them were restarted from saved progress during the run, as recorded in a dated amendment. %

\paragraph{Results.} On the eligible states (pooled ratio, with the wider of block and crossed-bootstrap intervals), KQ changed 6.7\% [0.8, 14.4] of outcomes and the BF16 action 9.0\% [2.9, 18.2]. KQ had 10 rescues and 10 damages and the BF16 action had 10 and 17, and the flips coincided on 5 rescues and 7 damages. Rescue and damage shares (Figure~\ref{fig:refs}b) were 3.3\% [0.0, 9.3] and 3.3\% [0.4, 8.0] for KQ, and 3.3\% [0.0, 9.2] and 5.7\% [1.5, 11.5] for the BF16 action. Over the full pool of 1700 units, KQ contributed 1.2\pp{} [0.1, 2.8] of turnover, against 17.9\pp{} for the BF16 action swap on all divergent states. %
\section{Selecting Decisions for Signed Benefit}
\label{app:r3}

This appendix supports \S\ref{sec:r4}. It gives the development diagnostics behind the frozen trigger, its secondary validation on BF16 histories, the matched random and hindsight references, and the exact random-selection test.

\subsection{Development diagnostics and secondary validation}
\label{app:token}
On the first development block, we scored each ordinary state at the token level (1140 configuration $\times$ state rows, 58 outcome changes), teacher-forcing both models on the generated token ids of $a_{\mathrm{B}}$ and $a_{\mathrm{Q}}$. We recorded the entropy truncated to the top 20 tokens, the top-1 margin and a Kullback--Leibler (KL) divergence over the union of both top-20 sets, in place of full entropy and KL. These analyses are exploratory and use development data. %

\paragraph{Uncertainty and divergence.} The quantized model's entropy at the start of its action separated states where the BF16 and quantized models act differently from states where they act identically, with an area under the receiver operating characteristic curve (AUC) of 0.66--0.91 per state source and model. Over all ordinary states from the quantized model's own histories, outcome changes also sat at lower quantized-model margins (AUC 0.31 [0.22, 0.40] for Qwen3-4B and 0.24 [0.13, 0.36] for Qwen3-8B, stratified). %
Among states that were already divergent, the AUC interval of every statistic visible to the quantized model covered 0.5, and associations in small subsets varied across state sources. This diagnosis motivated the frozen margin rule, whose validation reused the development tasks with new prompt blocks. %

\paragraph{Secondary validation on BF16 histories.} The secondary validation population consists of ordinary states from BF16 trajectories in the two later blocks (1020 rows, 75 outcome-changing states). At the 5, 10 and 20\% rates the frozen rule captured 4, 11 and 35 outcome-changing states, against 3.81, 8.35 and 16.07 expected at random. The excess was unresolved at 5 and 10\% ($p=.5404$ and $.1995$) and resolved at 20\%. The net gains were $-2$, $-1$ and $+5$, against $+0.07$, $+1.10$ and $+2.48$ at random. %

\subsection{Matched references for the frozen trigger}
\label{app:oracle}

Table~\ref{tab:oracle} compares the trigger of \S\ref{sec:r4} with matched random selection and a hindsight oracle on the same 1020 validation rows (states from the quantized model's own histories, two later prompt blocks), which contain 38 rescues and 31 damages. The \emph{matched action-replacement oracle} selects, per configuration, exactly as many rows as the frozen rule did. It takes only rows where action replacement rescues the outcome ($G>0$) and fills with unchanged rows ($G=0$) when a configuration has too few rescues, so it selects no damages. It is a hindsight bound rather than a deployable policy. At the 10\% and 20\% rates every rescue fits within the selected count. %
In Table~\ref{tab:oracle}B, the frozen rule's net gain at the 5\% rate exceeded matched random selection at the nominal 5\% level (exact one-sided $p=.0345$). After Holm correction across the three rates this became $p=.1035$, and the net gain stayed unresolved at every rate. Of the oracle's rescues, the frozen rule selected 6, 7 and 16 at the three rates. %

\begin{table*}[t]
\centering%
\renewcommand{\arraystretch}{1.12}\fontsize{8.5}{10}\selectfont\setlength{\tabcolsep}{4pt}
\begin{tabular*}{\textwidth}{@{\extracolsep{\fill}}lllll@{}}
\toprule
\multicolumn{5}{@{}l}{\textbf{A. Frozen trigger and matched hindsight oracle}} \\
\addlinespace[2pt]
Rate & sel. & oracle r/d/net (pp) & frozen r/d/net (pp) & net ratio \\
\midrule
5\% & 50 & 34/0/$+34$ ($+3.33$) & 6/2/$+4$ ($+0.39$) & 4/34 \\
10\% & 114 & 38/0/$+38$ ($+3.73$) & 7/5/$+2$ ($+0.20$) & 2/38 \\
20\% & 222 & 38/0/$+38$ ($+3.73$) & 16/11/$+5$ ($+0.49$) & 5/38 \\
\bottomrule
\end{tabular*}
\par\vspace{6pt}
\begin{tabular*}{\textwidth}{@{\extracolsep{\fill}}lllll@{}}
\toprule
\multicolumn{5}{@{}l}{\textbf{B. Frozen trigger against matched-count random selection (Figure~\ref{fig:trigger})}} \\
\addlinespace[2pt]
Rate & sel. & net vs.\ random ($p$; Holm) & r/d & turnover vs.\ random ($p$) \\
\midrule
5\% & 50 & $+4$ vs.\ $+0.21$ (.0345; .1035) & 6/2 & 8 vs.\ 3.48 (.0182) \\
10\% & 114 & $+2$ vs.\ $+0.98$ (.4169; .5234) & 7/5 & 12 vs.\ 7.52 (.0613) \\
20\% & 222 & $+5$ vs.\ $+2.38$ (.2617; .5234) & 16/11 & 27 vs.\ 14.30 (.0002) \\
\bottomrule
\end{tabular*}
\par\vspace{6pt}
\caption{Frozen trigger, which reads only the quantized model, against a matched action-replacement oracle (hindsight, same per-configuration selection counts) and matched-count random selection, on the 1020 validation rows from the quantized model's own histories. Rate: nominal selection rate. sel.: number of selected states. r/d/net: rescues, damages and net gain, with the net gain in percentage points (pp) of all rows in parentheses. Net ratio: frozen net gain over oracle net gain. Random: matched-count per-configuration random-selection expectation. $p$: the exact one-sided random-selection tail defined below. Holm correction across the three rates applies to the net tests only.}
\label{tab:oracle}
\end{table*}

\paragraph{Exact random-selection test.} In configuration $c$, condition on its $N_c$ states, containing $R_c$ rescues, $D_c$ damages and $Z_c=N_c-R_c-D_c$ unchanged outcomes. Draw $m_c$ states uniformly without replacement, matching the trigger's actual count. The random rescue and damage counts have joint mass
\[
 P_c(r,d)=\frac{\binom{R_c}{r}\binom{D_c}{d}\binom{Z_c}{m_c-r-d}}{\binom{N_c}{m_c}}.
\]
We convolve these distributions across configurations. The net statistic is $T=\sum_c(r_c-d_c)$, and its one-sided $p$-value is $P(T\ge T_{\rm trigger})$. The capture statistic uses $\sum_c(r_c+d_c)$ instead. The test accounts jointly for rescues and damages. It conditions on the observed validation rows, so its comparison with random selection applies to those rows rather than to new tasks.

\end{document}